\documentclass[a4paper]{styles/svproc}

\usepackage{graphicx}
\usepackage[noend]{algpseudocode}
\usepackage{amsmath,amssymb,amsfonts}
\usepackage{wrapfig}
\usepackage{algorithmicx, algorithm}
\usepackage{textcomp}
\usepackage[dvipsnames]{xcolor}
\usepackage{soul}
\usepackage{cite}

\usepackage{array}
\usepackage{booktabs}
\usepackage{tikz}
\usepackage[export]{adjustbox}
\usepackage{multirow}
\usepackage{siunitx}
\DeclareSIUnit{\rad}{rad}
\usepackage{url}

\usepackage[normalem]{ulem}

\usepackage[breaklinks,colorlinks]{hyperref} %
\usepackage[capitalize]{cleveref}

\newcommand\tab[1]{Table~\ref{#1}}
\newcommand\fig[1]{Fig.~\ref{#1}}

\long\def\invis#1{}

\DeclareMathOperator*{\argmax}{arg\,max}

\newcommand\copyrighttext{%
  \footnotesize \textcopyright 2022 Springer. Personal use of this material is permitted.
  Permission from Springer must be obtained for all other uses, in any current or future
  media, including reprinting/republishing this material for advertising or promotional
  purposes, creating new collective works, for resale or redistribution to servers or
  lists, or reuse of any copyrighted component of this work in other works.\\
  Presented at the International Symposium on Robotics Research (ISRR) 2022.\\
  DOI: TBA}
\newcommand\copyrightnotice{%
\begin{tikzpicture}[remember picture,overlay]
\node[anchor=south,yshift=20pt] at (current page.south) {\fbox{\parbox{\dimexpr\textwidth-\fboxsep-\fboxrule\relax}{\copyrighttext}}};
\end{tikzpicture}%
}

\begin{document}
\mainmatter

\title{Monocular Camera and Single-Beam Sonar-Based Underwater Collision-Free Navigation with Domain Randomization}
\titlerunning{Collision-Free Navigation}

\author{Pengzhi Yang\inst{1}\thanks{Authors contributed equally.} \and Haowen Liu\inst{2}$^\star$ \and Monika Roznere\inst{2} \and Alberto Quattrini Li\inst{2}}
\authorrunning{Yang et al.}

\tocauthor{Pengzhi Yang, Haowen Liu, Monika Roznere, Alberto Quattrini Li}

\institute{University of Electronic Science and Technology of China, Sichuan, China, 610056\\
{\tt\small peyang@alu.uestc.edu.cn}
\and
Dartmouth College, Hanover, NH, USA, 03755,\\
{\tt\small \{haowen.liu.gr, monika.roznere.gr, alberto.quattrini.li\}@dartmouth.edu}
}

\maketitle
\copyrightnotice
\begin{abstract}
Underwater navigation presents several challenges, including unstructured unknown environments, lack of reliable localization systems (e.g., GPS), and poor visibility.
Furthermore, good-quality obstacle detection sensors for underwater robots are scant and costly; and many sensors like RGB-D cameras and LiDAR only work in-air.
To enable reliable mapless underwater navigation despite these challenges, we propose a low-cost end-to-end navigation system, based on a monocular camera and a fixed single-beam echo-sounder, that efficiently navigates an underwater robot to waypoints while avoiding nearby obstacles.
Our proposed method is based on Proximal Policy Optimization (PPO), which takes as input current relative goal information, estimated depth images, echo-sounder readings, and previous executed actions, and outputs 3D robot actions in a normalized scale.
End-to-end training was done in simulation, where we adopted domain randomization (varying underwater conditions and visibility) to learn a robust policy against noise and changes in visibility conditions.
The experiments in simulation and real-world demonstrated that our proposed method is successful and resilient in navigating a low-cost underwater robot in unknown underwater environments. The implementation is made publicly available at {\scriptsize \url{https://github.com/dartmouthrobotics/deeprl-uw-robot-navigation}}.
\end{abstract}


\section{Introduction}

This paper presents an integrated deep-learning-based system, contingent on monocular images and fixed single-beam echo-sounder (SBES) measurements, for navigating an underwater robot in unknown 3D environments with obstacles.

Obstacle avoidance is fundamental for Autonomous Underwater Vehicles (AUVs) to safely explore the largely unmapped underwater realms (e.g., coral reefs, shipwrecks).
However, the underwater environment itself poses unique challenges in regards to safe navigation, which is still an open problem for AUVs~\cite{petillot2019underwater}.
There are limited sensors and positioning systems (e.g., GPS) that accurately measure the surroundings and operate underwater, thus preventing the use of well-established navigation methods~\cite{pfrunder2017real} that were originally designed for ground vehicles with sensors like LiDAR.
In addition, the sensor configurations in low-cost AUVs, equipped with monocular camera, inexpensive IMU, compass, and fixed SBES, bear their own individual drawbacks, such as no scale information and drifting/uncertain measurements.
These challenges make the classic methods for obstacle avoidance and navigation in unknown environments -- i.e., those which (1) estimate the geometry of the space using sensors with direct~\cite{engel2017direct} or indirect~\cite{campos2021orb, rahman2019iros-svin2} state estimation methods and (2) apply specific behaviors or planning in the partial map (e.g., Vector Field Histogram~\cite{Panagou2014}, Dynamic Window Approach~\cite{fox1997dynamic}%
) --  not directly applicable in underwater scenarios.

\begin{wrapfigure}[19]{R}{0.48\textwidth}
\vspace{-2.5em}
\includegraphics[width=0.48\textwidth]{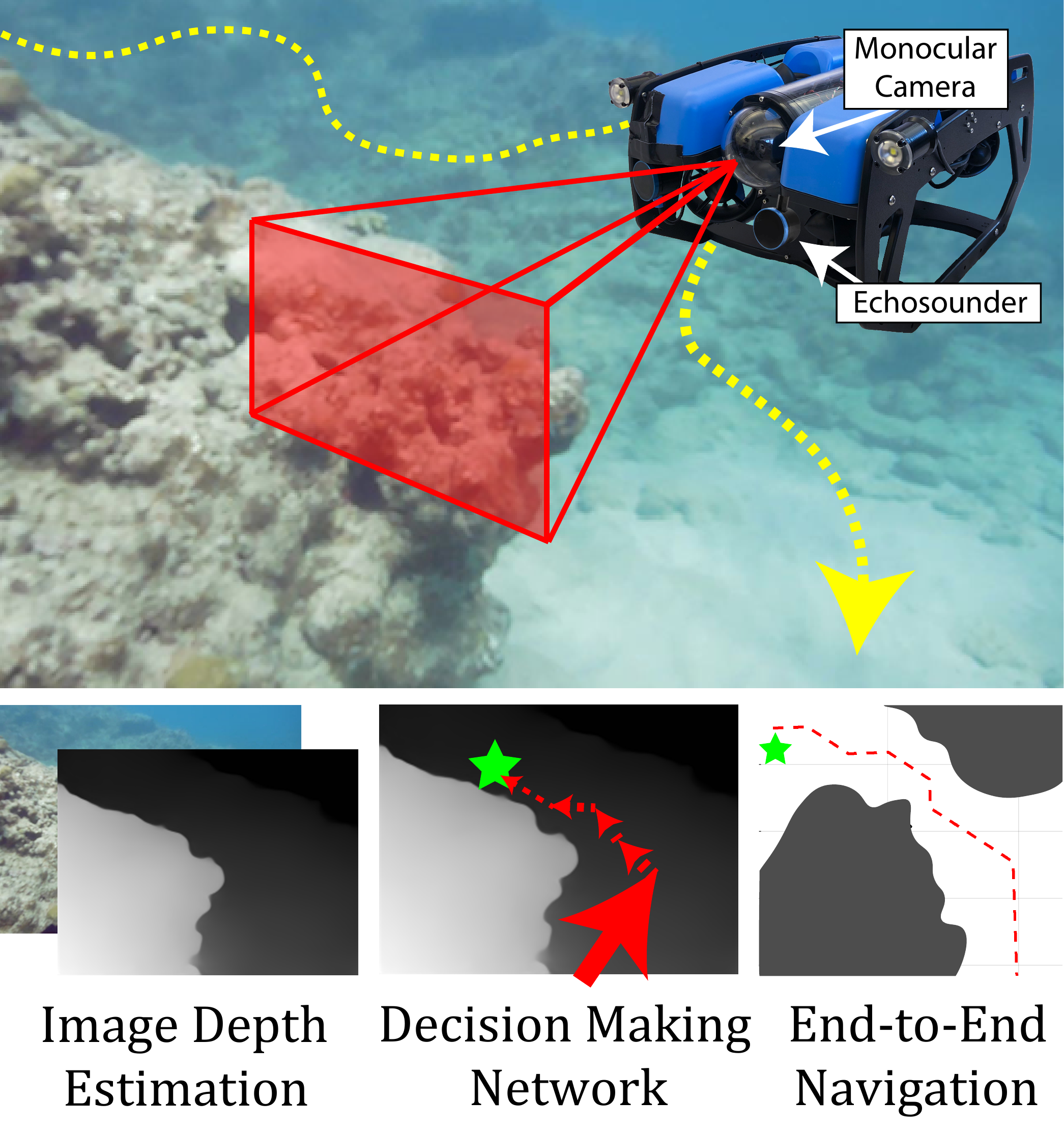}\vspace{-1.2em}
\caption{How to guide an underwater robot to 3D waypoints given only monocular images, fixed echo-sounder range measurements, and a localization system, but \emph{no map}, while also avoiding obstacles?}
\label{fig:beauty}
\end{wrapfigure}

With recent advances in deep reinforcement learning (DRL)\\~\cite{kober2013reinforcement,%
mnih2015human}, several end-to-end deep neural network based methods have emerged, from raw images to control outputs.
These end-to-end methods -- typically tasked to endlessly navigate or reach a visual target -- demonstrated good performance for ground robots in unknown environments~\cite{xie2018wheels%
}. Comparatively, underwater domains bring problems to learning-based vision navigation due to a more complex image formation model that results in, e.g., backscattering and light attenuation. %

This paper proposes a goal-oriented end-to-end DRL navigation approach, given that classical planning methods are not straightforward to apply as they require accurate maps, which are difficult to obtain due to the underwater perception challenges described above.  %
In particular, we design the first multi-modal end-to-end underwater navigation system in unstructured 3D  environments for which no map is available, based on Proximal Policy Optimization (PPO) \cite{wijmans2019ddppo}, which allows for continuous action space. The provided inputs are goal positions, estimated depth images, and range measurements from the fixed SBES. Monocular camera and fixed SBES keep the AUV's cost low, while exploiting and complementing the individual sensor' strengths -- i.e.,  large field of view from the monocular camera that can provide relative scene depth and the absolute range measurement from the SBES. %
We also propose a method to mitigate the sim-to-real gap problem by leveraging domain randomization into our system. We generated realistic simulated environments with different underwater visibility and randomized training environments, enhancing the model robustness to the changing visual conditions in real underwater domain.
    Extensive experimental analysis with tests and ablation studies of the proposed navigation system were conducted both in simulation and real-world. Results demonstrated high safety and efficiency compared to traditional navigation baselines and other sensor/model configurations, as well as reliable transferability to new environments.
\section{Related Work}\label{sec:relatedwork}
Obstacle avoidance and navigation without a prior map has been studied starting with wheeled mobile robots equipped with bumpers and sonar sensors~\cite{choset2005principles} and later branching off into different environments and sensor configurations.
For underwater domains, one of the main challenges is the limit of choices for sensors.
While some underwater LiDAR solutions are available~\cite{mcleod2013autonomous}, they are expensive (US\$100,000 or more) and bulky -- requiring a laser scanner and a camera. In addition, there is a lack of global positioning systems and the acoustic based positioning systems are affected by noise, making mapping    underwater challenging~\cite{petillot2019underwater}.
Our goal is to enable navigation for low-cost AUVs. Therefore, in the following, we discuss applications using sensors (i.e., SBES, cameras) that are typically configured on low-cost underwater robots.

In practice, many underwater navigation systems depend on acoustic, inertial, and magnetic sensors \cite{kinsey2006navigation,williams2001navigation,paull2013auv}.
For example, Calado \textit{et al.}~\cite{calado2011obstacle} proposed a method where the robot used a SBES to detect obstacles and construct a map of them.
However, SBES can only provide a fixed single distance measurement and has high uncertainty given the wide beam cone -- around \ang{30}. 
To infer more about the complex scene, the robot must frequently turn in multiple directions, which negatively affects navigation efficiency. 
Alternatively, multi-beam and mechanical scanning sonars can cover a larger field of view~\cite{petillot2001underwater}. 
Hern\'{a}ndez \textit{et al.}~\cite{hernandez2015online} used a multi-beam sonar to simultaneously build an occupancy map of the environment and generate collision-free paths to the goals. Grefstad \textit{et al.}~\cite{grefstad2018navigation} proposed a navigation and collision avoidance method using a 
mechanically scanning sonar for obstacle detection.
However, a scanning sonar takes a few seconds to scan a 360$^{\circ}$ view. 
The acoustic sensors' accuracy depends on the environment structure and the type of reflections that arise. In addition, multi-beam and mechanical scanning sonars are significantly more expensive than monocular cameras and SBES (in the order of $>$US\$10k vs.\ US\$10 - US\$100).  

While cameras have shown to provide dense real-time information about the surroundings out of the water~\cite{liu2015learning}, there are fewer underwater obstacle avoidance methods that use cameras. The underwater domain indeed poses significant challenges, including light attenuation and scattering.  
Most work considers reactive controls, i.e., no goal is specified. 
Rodr{\'\i}guez-Teiles \textit{et al.}~\cite{rodriguez2014vision} segmented RGB images to determine the direction for escape. Drews-Jr \textit{et al.}~\cite{drews2016dark} estimated a relative depth using the underwater dark channel prior and used that estimated information to determine the action. 
There has been recent efforts in 3D trajectory optimization for underwater robots.
Xanthidis \textit{et al.}~\cite{xanthidis2020navigation} proposed a navigation framework for AUV planning in cases when a map is known or when a point cloud provided by a visual-inertial SLAM system~\cite{rahman2019iros-svin2} is available. Our proposed method navigates the robot to 3D waypoints without explicit representation of the environment.

Recently, deep learning (DL) methods have shown to work well with underwater robots.
Manderson \textit{et al.}~\cite{manderson2018vision} proposed a convolutional neural network that takes input RGB images and outputs unscaled, relative path changes for AUV driving. The network was trained with human-labeled data with each image associated with desired changes in yaw and/or pitch to avoid obstacles and explore interesting regions. 
Later it was extended with a conditional-learning based method for navigating to sparse waypoints, while covering informative trajectories and avoiding obstacles~\cite{manderson2020vision}. Our proposed method does not require human-labeled data.

Amidst the progress in DRL, there is more research on robots operating out of water with monocular cameras.
Some of these methods addressed the problem of safe endless 2D navigation without specifying any target location. 
Xie \textit{et al.}~\cite{xie2017monocular} trained a Double Deep Q-network to avoid obstacles in simulated worlds and tested it on a wheeled robot. Kahn \textit{et al.}~\cite{kahn2018self} proposed a generalized computation graph for robot navigation that can be trained with fewer samples by subsuming value-based model-free and model-based learning. 
Other works provided the goal as a target image instead of a location~\cite{zhu2017target,devo2020towards,wu2020towards}. %
Some methods, based on an end-to-end network, guided the robot to the goal using LiDAR or RGB-D cameras~\cite{pfeiffer2017perception,%
xie2018wheels,%
zhang2017deep, liang2021crowd} and goal's relative position for path planning. %
Recently, a DD-PPO based method was used to navigate a robot in an unknown indoor (simulated) environment, using a RGB-D camera, GPS, and compass~\cite{wijmans2019ddppo}. Our method will be based on PPO, with the additional challenge of not having  depth information directly from the camera.

Nevertheless, due to the difficulties of applying DRL in real-world environments, most works performed training in simulation.
However, policies learned in simulated environments may not transfer well to the real-world environment, due to the existence of reality (sim-to-real) gap~\cite{tobin2017domain}.
To address this, several methods utilized domain randomization, where parameters of the simulated world were varied so that policies learned remained robust in real-world domain.
For example, Sadeghi and Levine~\cite{sadeghi2016cad2rl} proposed a DRL approach for indoor flight collision avoidance trained only in CAD simulation that was able to generalize to the real world by highly randomizing the simulator's rendering settings. %

Our approach draws from the advances in DRL: we design an end-to-end pipeline for low-cost underwater robot navigation to address the underwater challenges, combining multiple sensors and applying domain randomization.
\begin{figure*}
    \centering
    \includegraphics[trim={0cm .4cm 0cm 0cm}, clip, width=.9\textwidth]{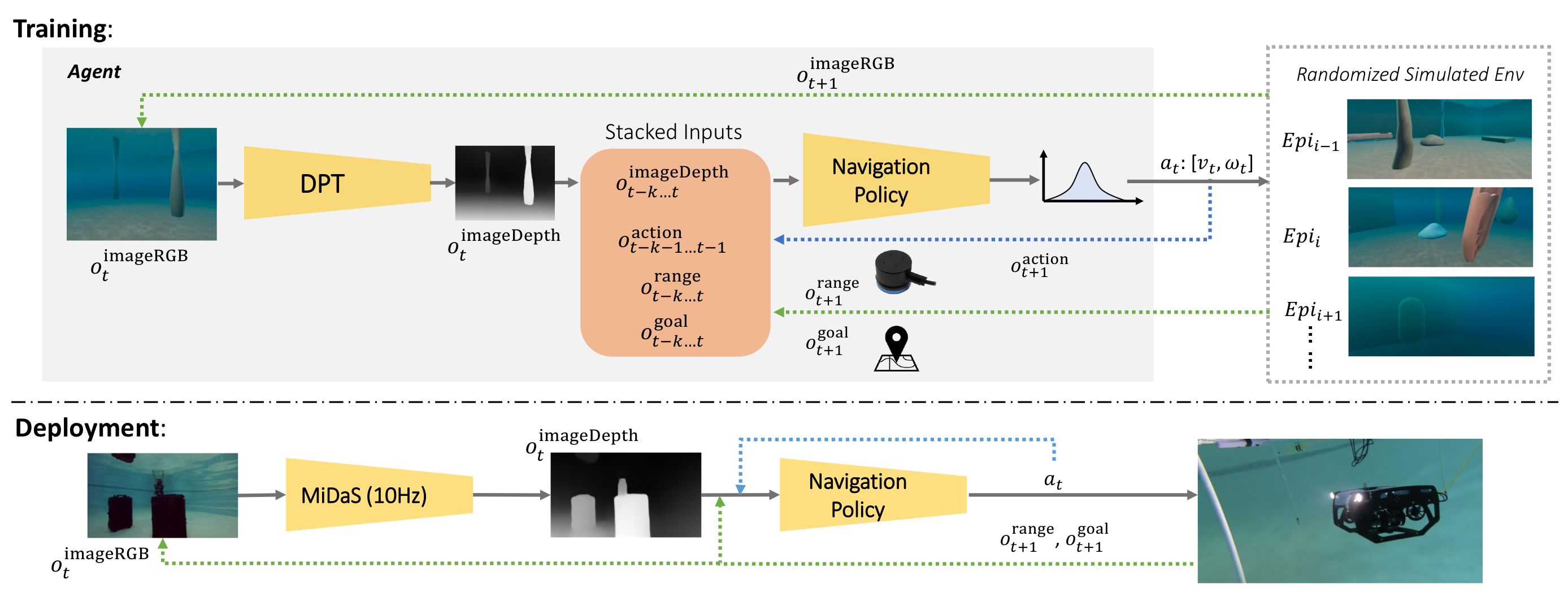}
    \vspace{-1em}
    \caption{\textit{Flowchart for the Proposed End-to-End Underwater 3D Navigation System.}
    The pipeline includes two stages: a depth prediction module (DPT) followed by a decision making module (PPO). During training, at each episode $i$, the robot is deployed in a randomized simulated environment. Predicted depth map $o^\textrm{imageDepth}_t$ of the raw RGB image $o^\textrm{imageRGB}_t$, relative goal position $o^\textrm{goal}_t$, echo-sounder reading $o^\textrm{range}_t$, and previous executed action $a_{t-1}$ are stacked with past $k$ observations from the previous times steps to feed into the PPO network (solid lines). The robot performs the action sampled from the output policy distribution. New observations (dashed lines) are then obtained for computing the next action at time step $t+1$. %
    During real-world deployment, DPT's computationally less expensive counterpart MiDaS was used as the depth prediction module for real-time inference.
    }
    \label{fig:system_overview}
    \vspace{-15pt}
\end{figure*}

\section{Approach}\label{sec:approach}
The problem considered in this paper is as follows: an underwater robot deployed in an unknown environment needs to navigate to a goal location $G \in \mathbb{R}^3$, minimizing the travel time, while avoiding collisions with obstacles. 

To develop a mapless navigation solution for low-cost robots, we consider an underwater thruster-vectored robot that has an inexpensive sensor suite composed of: (1) a monocular camera, (2) a SBES placed below the camera and looking forward, (3) a compass, (4) pressure sensor for water depth, and (5) a (noisy) localization system. Selecting this sensor configuration allows us to exploit the larger field of view (FOV) covered by the camera while obtaining absolute front distance estimates with the fixed SBES.

For a general solution, robust to noise and changing visual conditions, we approach the real-time 3D navigation problem by devising an end-to-end system (  see \fig{fig:system_overview} ) based on a neural network for dense depth prediction from monocular images and on a deep reinforcement learning method that takes as input %
the sensor suite data and outputs vertical and steering commands. %
We consider a window of prior measurements and executed actions given the absence of prior knowledge of the environment. 

In the remainder of this section, we describe in detail the RL approach, the depth prediction network, and how to address the sim-to-real gap.

\subsection{Multi-Modal Deep Reinforcement Learning Navigation}
Given an unknown environment, the navigation problem can be formulated as a Partially Observable Markov Decision Process (POMDP), defined with a 6-tuple: state space $S$ that cannot be directly observed by the robot, action space $A$ modifying the current state of the robot, observation space $\Omega$, a state-transition model $T$, the observation probability distribution $O$, and a reward function $R$ which returns the reward after a state transition.

\textbf{Observation space.} The observation $O_t$ at time step $t$ consists of: (1) the predicted depth image $o^\textrm{imageDepth}_t \in \mathbb{R}^{128\times160}$; (2) an SBES range measurement $o^\textrm{range}_t \in \mathbb{R}$; (3) the current relative goal position $o^\textrm{goal}_t \in \mathbb{R}^3$ -- specifically, $[D^h_t, D^v_t, \theta^h_t]^\top$, where $D^h_t$, $D^v_t$ are robot's current horizontal, vertical distances to the goal and $\theta^h_t$ represents the relative yaw heading difference; and (4) the past executed actions $o^\textrm{action}_t \in \mathbb{R}^2$. We stack observations considering a time window $k$ to capture the robot's progress towards the goal and to avoid obstacles that left the periphery view. In experiments, model using 5 time steps (decision period lasts 0.5 second for each step) showed good performance without adding too much computational expense.

\textbf{Action space.} The action space is $a_t = [v_t,\omega_t] \in \mathbb{R}^2$, where $v_t$ is the vertical linear velocity and $\omega_t$ is the yaw angular velocity. To generalize the applicability of the learned behavior to different robots, we consider the actions to be in a range of $[-1.0, 1.0]$ which will be linearly mapped to the range of velocities of a specific robot. Note that while we could include the horizontal forward linear velocity, we decided to keep it constant to facilitate surveying missions that require the same velocity to collect consistent high-quality measurements. 

The action is then given by the policy:
\begin{equation}
\small
    a_t = \pi(O_t)%
\end{equation}
The goal is to find the optimal policy $\pi^*$ which maximizes the navigation policy's expected return over a sequence $\tau$ of observations, actions, and rewards:
\begin{equation}
\small
    \pi^* = \argmax_\pi \mathbb{E}_{r\sim p(\tau|\pi)}\Big[\sum\gamma^t r_t\Big]
\end{equation}
\noindent where $\gamma \in [0,1.0]$ is the discount factor. The optimal policy would translate in a path that is safe and minimizes the time it takes to travel to the goal.

\textbf{Reward function.} Our reward function $r_t$ at time $t$ encodes the objectives to stay not too close to any obstacle ($r^{\textrm{obs}}_t$) and to reach the goal area as soon as possible ($r^{\textrm{goal}}_t$). 

When the robot is close to an obstacle, it will compute a negative reward: %
\begin{equation}
\small
    r_t^{\textrm{obs}} = 
    \left\{
        \begin{array}{lr}

         -r_{\textrm{crash}}, & d_t^h < \delta_h \lor\,d_t^v < 
            \delta_v  \lor\,d_t^{\textrm{sur}} < \delta_v\\
    -s_0(2\delta_h - d_t^h), & \delta_h \leq d_t^h < 2\delta_h \\
            0 & \textrm{otherwise}
        \end{array}
    \right.
\end{equation}
where $\delta_h$, $\delta_v$ represent the thresholds for the distances of the robot to the closest obstacle $d_t^h$, $d_t^v$ -- horizontally or vertically, respectively. 
We also check the distance to the water surface $d_t^{\textrm{sur}}$, as there might be surface obstacles that cannot be detected given the sensor configuration of the robot.
The threshold values $\delta_h$, $\delta_v$ should consider the robot's size and turning radius.
When any of the constraints are met -- i.e., the robot is too close to an obstacle or the surface -- the current episode terminates with a large negative constant reward $-r_{\textrm{crash}}$.
In addition, to guarantee safety, a penalty for motions within a range $[\delta_h, 2\delta_h)$ of distance to nearby obstacles is given according to the current distance.
Otherwise, if the robot is far from the obstacles, no negative reward is applied.

To guide the robot towards the goal both horizontally and vertically, 
we split the goal-based reward into two parts.
First, the horizontal goal-based reward:
\begin{equation}
\small
    r_t^{\textrm{goalh}}  = 
    \left\{
        \begin{array}{lr}
             -s_1|\theta_t^h|, & \Delta_h < D_t^{h} \\
            r_{\textrm{success}} - s_2|\theta_t^h|, & \textrm{otherwise}%
        \end{array}
    \right.
\end{equation}
If the robot's horizontal distance to the goal $D_h^t$ is greater than a threshold $\Delta_h$, 
then the penalty is based on the robot's orientation to the goal -- i.e., a robot already facing the goal gets a smaller penalty, as the constant forward velocity will ensure shorter arrival time.
Otherwise, if the robot is within the goal area, then there is a positive reward with a preference to the robot's orientation towards the goal.

Likewise, the vertical goal-based reward:
\begin{equation}
\small
    r_t^{\textrm{goalv}}  = 
    \left\{
        \begin{array}{lr}
            s_3|\dot{D}_t^v|%
            , &  \dot{D}_t^v \leq 0 \land\,\Delta_h < D_t^h \\
            - s_3|\dot{D}_t^v|%
            , &  \dot{D}_t^v > 0 \land\,\Delta_h < D_t^h \\
            - s_4|D_t^v|, & \textrm{otherwise}%
        \end{array}
    \right.
\end{equation}
When the robot is not near the goal, the vertical goal-based reward is a positive value if the change in vertical distance over time $\dot{D}_t^v$ is negative or 0 -- i.e., the robot is getting closer to the target depth.
On the contrary, it is a negative value if the change is positive -- i.e., the robot is getting farther from the target depth.
Otherwise, if the robot is within goal area, the negative reward is relative to the distance to the target depth.
This split (horizontal and vertical) of the goal reward showed better stability in experiments than when a single combined goal reward was applied, potentially due to the separate focus of two mostly independent actions.

The above obstacle- and goal-based rewards conflict with each other; they could lead to oscillations at local optima when an obstacle is nearby.
Thus, we devised a priority-based strategy (when the robot is not in the goal area) that focuses on moving away from the obstacle by scaling $r_t^{\textrm{goalh}}$:
\begin{equation}
\small
    \begin{array}{lr}
        r_t^{\textrm{goalh}} \ *\!= 
        s_5(d_t^h - \delta_h)/\delta_h, &  \Delta_h < D_t^{h} \land \delta_h \leq d_t^h < 2\delta_h \label{con:priority}
    \end{array}
\end{equation}

In all the reward equations, $s_0, \ldots ,s_5$ are positive scaling factors. Intuitively, they are set so that rewards are in an appropriate scale for a balanced training performance. 

Finally, the collective reward at time $t$ can be obtained as:
\begin{equation}
\small
    r_t = r^{\textrm{obs}}_t + r^{\textrm{goalh}}_t + r^{\textrm{goalv}}_t
\end{equation}

\begin{figure}
   \centering
    \includegraphics[trim={0cm 0.cm 0.cm 0},clip,width=0.45\textwidth]{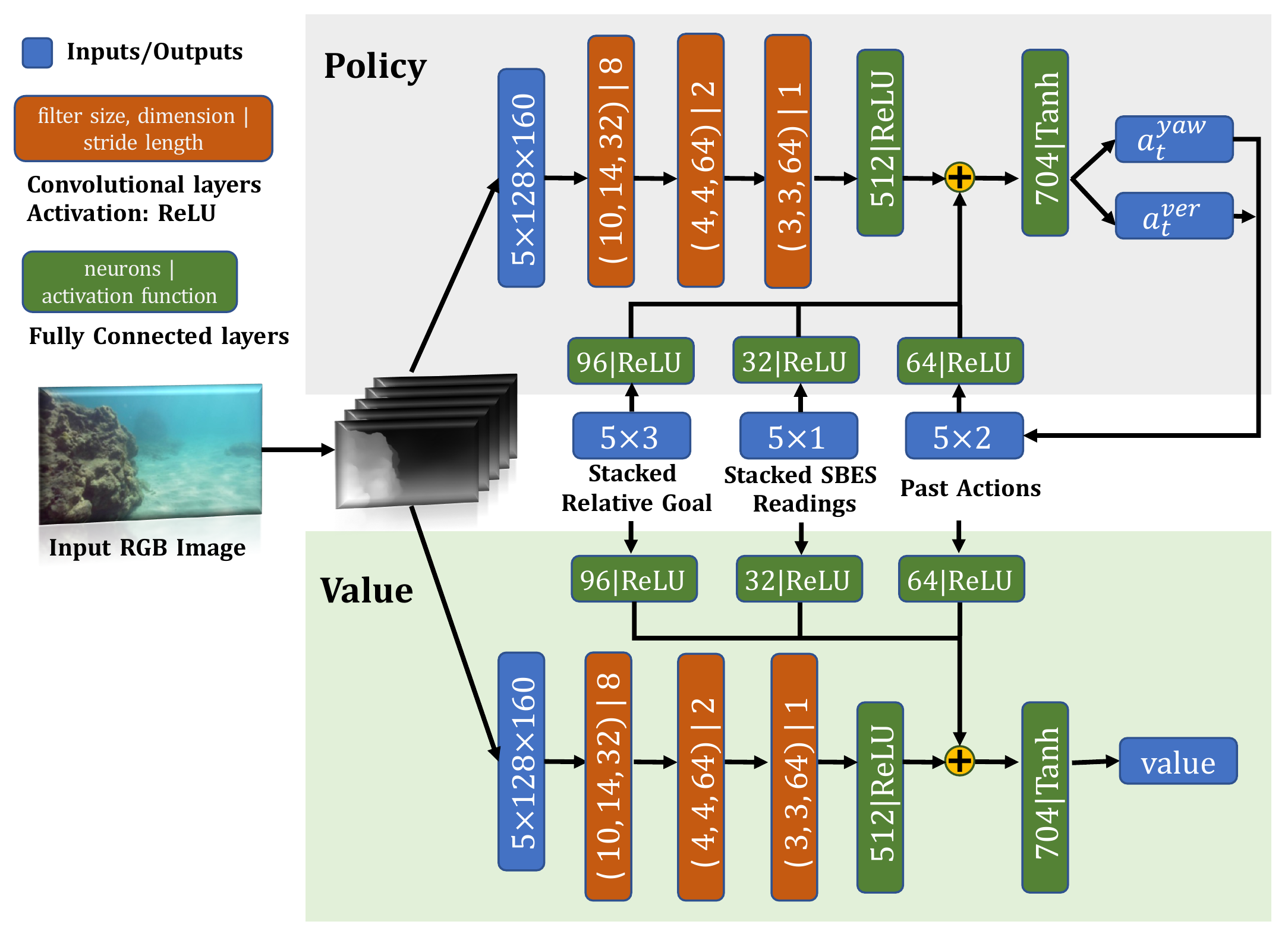}
    \includegraphics[width=0.4\textwidth]{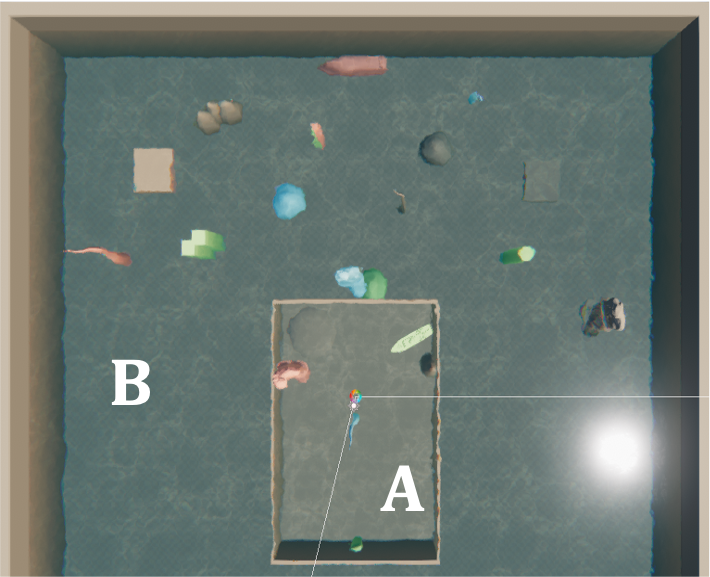}
    \vspace{-1em}
    \caption{(left) \textit{Network Architecture.} Predicted depth images are processed by three layers of convolutional layers (orange). Its output is flattened and concatenated with feature vectors (green) representing the stacked relative goal positions, echo-sounder readings, and past actions. The final fully-connected layer outputs a navigation policy and state value. (right) \textit{Top View of the Training Env.} Our model was trained in the above simulated environment in area A (inside area with fewer obstacles and smaller space) and B (outside area with more obstacles and larger space).}
    \label{fig:training_setting}
    \vspace{-2em}
\end{figure}

\textbf{Network architecture.}
The network structure depicted in \fig{fig:training_setting}(left) illustrates how we integrate the information vectors from the sensors. First, the stacked predicted depth images are processed by three convolutional layers, then the flattened output $\in \mathbb{R}^{512}$ is concatenated with processed feature vectors consisting of the stacked relative goal positions $\in \mathbb{R}^{96}$, SBES readings $\in \mathbb{R}^{32}$, and past actions $\in \mathbb{R}^{64}$. Specifically, the combined echo-sounder readings provide an implicit scale on the relative depth prediction without requiring calibration. The network will produce a navigation policy and state value. 

\subsection{Image Depth Prediction Network} \label{Combined Perception Inputs}
Accurate image depth predictions is important for our navigation pipeline to work.
Previous work used ground truth simulated depth images with Gaussian noise as input for training and applied depth estimation during deployment~\cite{xie2017monocular}.
However, this broadens the sim-to-real gap as real-world noise in depth predictions is more complex than implemented simulated noise models~\cite{sweeney2019supervised}.
Instead, we utilized one of the latest monocular depth prediction networks, Dense Prediction Transformer (DPT)~\cite{ranftl2021vision}, which has an encoder-decoder design and applies a transformer as the encoder's main building block. We selected DPT over other deep neural networks for depth prediction for its state-of-the-art performance in single-view depth estimation and robustness across diverse environments.  %

\subsection{Transferable Model} 
DRL often has the problem of generalization: models trained in one domain fail to transfer to other domains even if there are small differences between the domains~\cite{cobbe2019quantifying}. %
Unlike in-air, images taken underwater will look drastically different across various environments due to the more complex lighting and backscattering effects~\cite{akkaynak2018underwater}.
Thus, training the model in a single fixed environment would lead to over-fitting to that environment's visual conditions. %
One solution is to retrain the depth prediction network with an existing underwater image depth dataset, which, however, is not available. Another solution is to enhance the input underwater images to its approximate in-air counterpart~\cite{roznere2019color, %
akkaynak2018underwater}.
Yet, most image enhancement techniques require difficult-to-retrieve information (e.g., water attenuation coefficients, depth maps).

Our approach is to integrate underwater features into the simulation used for training.
We modified an existing underwater simulator framework for games to create the training and testing simulations for our proposed approach. The framework contains custom shaders that incorporates a light transmission model to simulate underwater optical effects, thus providing a good amount of realism.

\textbf{Domain randomization.} 
We integrated domain randomization to generate underwater environments with different visual conditions, thus enabling transferability. %
In particular, %
at the start of every training episode, we randomize the underwater visibility -- the gradient and conditions in visibility over distance.
Visibility was selected as it significantly impacts the relative depth estimation, thus affecting to a large extent how the robot perceives its surroundings.

We decided not to apply domain adaptation~\cite{peng2020learning} -- i.e., the process of learning different environment encoding and corresponding adapted policy during training, so that during testing the best environment encoding will be found with the corresponding adapted policy -- because searching the best environment encoding is not very practical for underwater deployments.
For instance, the search would require robot motions towards obstacles to identify the (potentially changing) visibility feature of the specific environment. %

\textbf{Multi-scenario training.}
We built the simulated training environment via Unity Engine\footnote{\scriptsize \url{http://www.unity.com/}}. %
We generated two activity areas to represent two classes of environments that an AUV might encounter: \textit{A} -- a small area with fewer obstacles, and \textit{B} -- a big cluttered area with obstacles at various positions and heights (see \fig{fig:training_setting}(right)).
In each training episode, the robot's starting pose and goal location are randomly reset in the environment.
This exposure to different training scenarios ensures that the learned policy will be more likely to handle more complex environments~\cite{ %
tobin2017domain}. %

\section{Experimental Results}
We trained and performed experiments in simulation, in real-world with a vector-thruster underwater robot, and with underwater datasets to validate our DRL-based multi-modal sensor navigation system.
We performed comparisons and ablation studies with other methods.
Our framework is publicly available\footnote{\scriptsize\url{https://github.com/dartmouthrobotics/deeprl-uw-robot-navigation}}.

\subsection{Training Experimental Settings}
Our model was first trained and tested on a workstation with two 12GB NVIDIA 2080Ti GPUs.
It was implemented with PyTorch and Adam optimizer~\cite{kingma2014adam}.

In simulation, the robot's forward velocity, vertical velocity range, and yaw angular velocity range were set to
\SI{0.345}{m/s}, 
\SIrange{-0.23}{0.23}{m/s},
\SIrange[parse-numbers = false]{-\text{$\pi$}/6}{\text{$\pi$}/6}{rad/s}, respectively.
While the training environment allows for higher velocities, we chose low velocities to avoid any ``jerky'' motion that could happen with the AUV at high speed. The camera's horizontal and vertical FOVs were set to \ang{80} and \ang{64}. The simulated echo-sounder's max detection range was set to \SI{4}{m}, which are all consistent with the real-world sensor configuration.
The simulation environments' visibility value was randomly chosen within the range of \SIrange{3}{39}{m}.

We trained for 250 iterations -- each with at least 2048 time steps -- and observed the reward was stable after around 120 iterations (learning rate of 3e-5).
The detailed constant and threshold values for the reward function -- i.e.,  $r_{\textrm{success}}$, $r_{\textrm{crash}}$, $\Delta_h$, $\delta_h$, and $\delta_v$ -- were set to $10$, $10$, \SI{0.6}{m}, \SI{0.5}{m} and \SI{0.3}{m}, while the scaling factors $s_0, s_1, \ldots, s_5$ were set to $2.0$, $0.1$, $1.0$, $1.0$, $8.0$, $1.0$.

\subsection{Performance Comparison with Different Sensor Configurations}
\vspace{-0.5em}
\begin{figure}[t]
  \centering
  \resizebox{\columnwidth}{!}{
  \begin{tabular}{cccc}
     \includegraphics[height=.5in, valign=c,trim={0cm 0.7cm 3.5cm 2.3cm},clip]{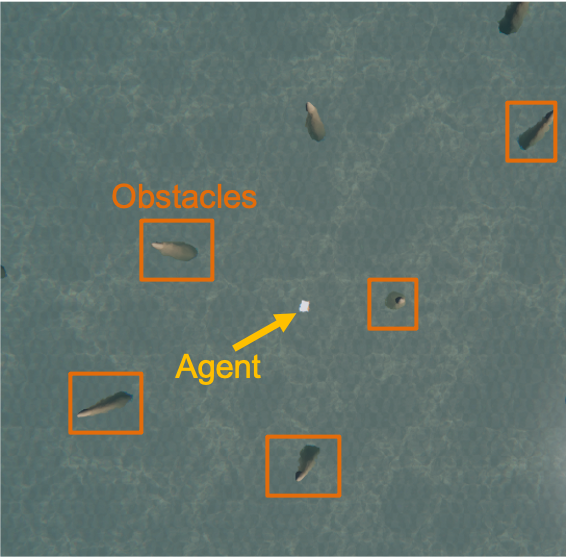}
     \includegraphics[height=.5in, valign=c,trim={4.6cm 0.7cm 4.95cm 1.3cm},clip]{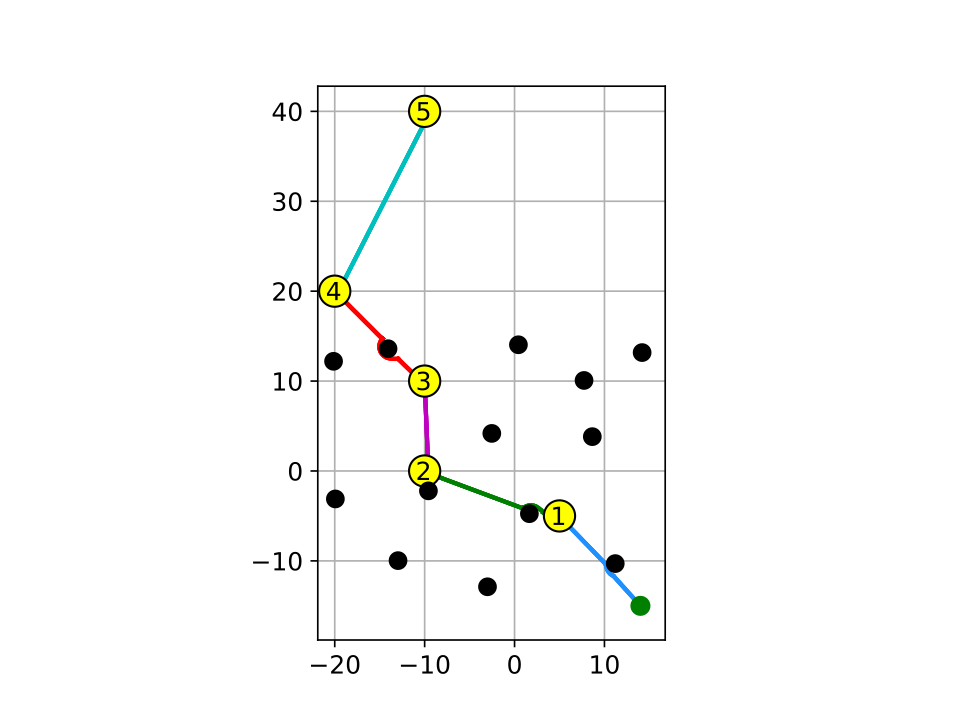}
     & \includegraphics[height=.5in, valign=c,trim={5.2cm 0.7cm 4.9cm 1.3cm},clip]{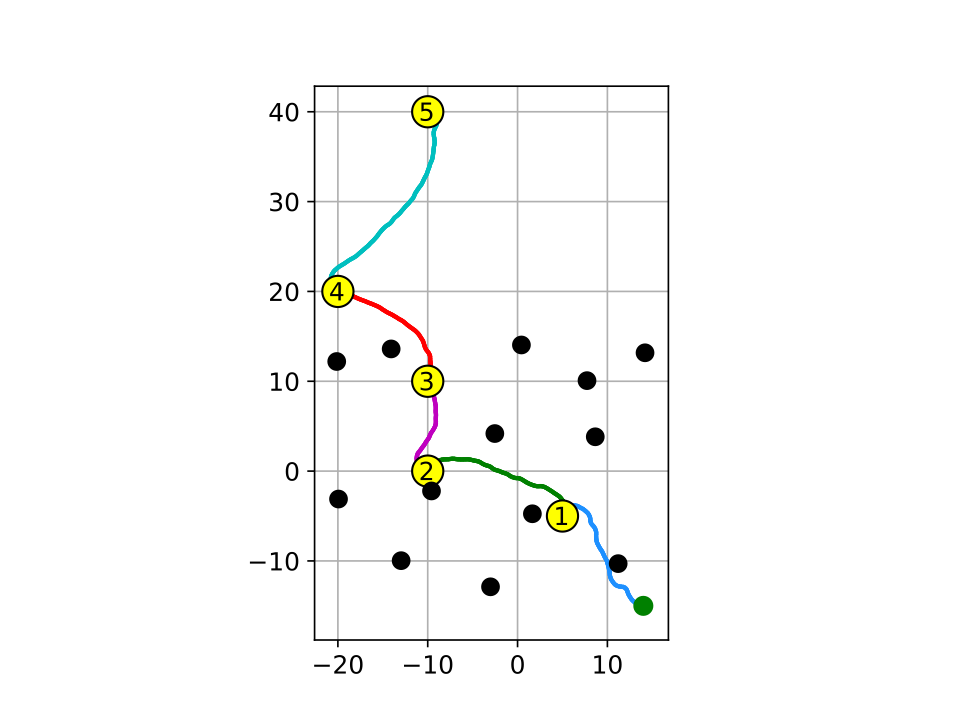}
     & \includegraphics[height=.5in, valign=c,trim={5.2cm 0.7cm 4.9cm 1.3cm},clip]{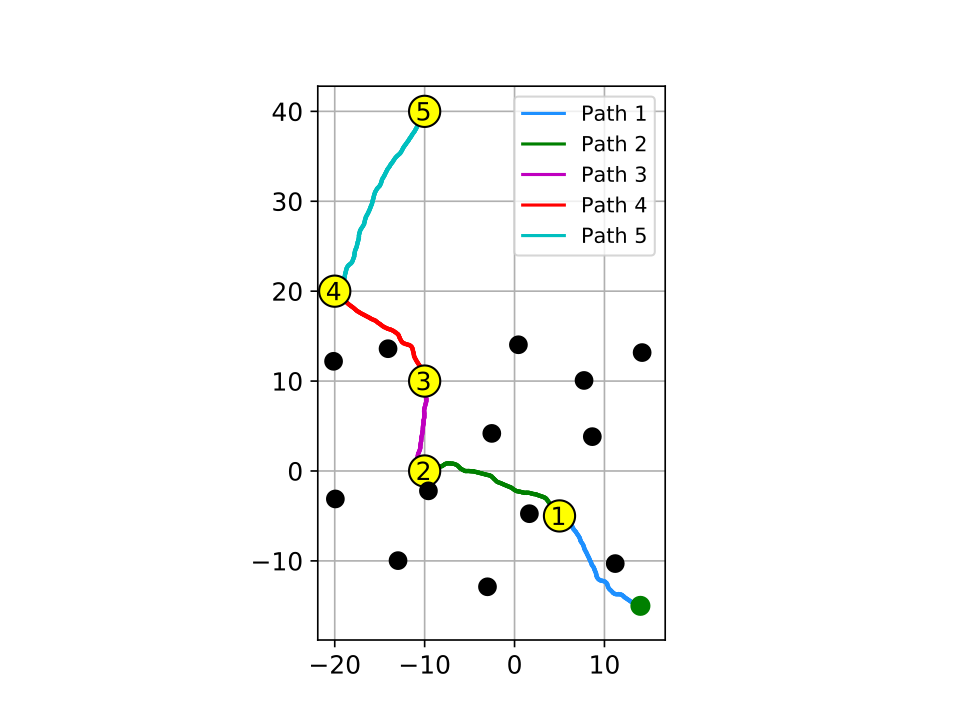} 
  \end{tabular}
  }
  \caption{\textit{Partial top view of runs in Cluttered Env. (left): Bug2 (second), Our Model w/o SBES (third), and Our Model w/ SBES (right).} Legend: robot's start pose (green dot); obstacles (black dots); waypoints to reach in order (circled numbers). %
  }
  \label{fig:Waypoint_tests_trajectories}
  \vspace{-2em}
\end{figure}

\begin{table*}[b]
\centering
\caption{\textit{Waypoint Tests Results.} 10 runs for each of the three methods: Bug2 with multi-beam sonar, our model trained without fixed single-beam echo-sounder, and our proposed model.
The travel time average and standard deviation (in seconds) of successful runs for each waypoint were calculated, as well as the overall success ratio to reach all five waypoints.
}
\label{Quantitative_Analysis_Waypoints}
\resizebox{\textwidth}{!}{
    \begin{tabular}{cccccccc}
        \toprule
        \multirow{2}*{Method} & \multirow{2}*{Sensors} & \multicolumn{5}{c}{Traveling Time/s (less is better)} & Success Ratio \\\cline{3-7}
           & & $wp1$ & $wp2$ & $wp3$ & $wp4$ & $wp5$ & (higher is better) \\
        \midrule
         Bug2 & MBS & 57.6 $\pm$ 0.3 & 66.95 $\pm$ 0.15 & 41.15 $\pm$ 0.45 & 69.8 $\pm$ 0.9 & 77.65 $\pm$ 0.45 & \textbf{100\%} \\
         Ours\ w/o\ SBES & Monocular Camera & 51.8 $\pm$ 5.94 & 56.5 $\pm$ 2.09 & 35.62 $\pm$ 8.07 & 47.0 $\pm$ 2.03 & 76.0 $\pm$ 2.21 & 40\% \\
         Ours\ w/ SBES & Monocular Camera \& SBES & \textbf{38.35 $\pm$ 0.45} & \textbf{49.8 $\pm$ 0.78} & \textbf{29.3 $\pm$ 0.78} & \textbf{44.3 $\pm$ 0.6} & \textbf{67.25 $\pm$ 0.6} & \textbf{100\%} \\\bottomrule
    \end{tabular}
}

\end{table*}

We first tested the efficiency of our proposed multi-modal low-cost navigation approach against a traditional metric-based goal-oriented navigation method that does not require any map, given that no map of the underwater environment is available. In particular, we selected Bug2 algorithm given its guarantees on the path length. To have Bug2 work effectively, we employed a multi-beam sonar (MBS), a common but expensive sensor for underwater obstacle avoidance, which emits multiple beams in a plane with a typical horizontal FOV of $120^{\circ}$. %
We also considered our model trained without the echo-sounder as ablation study to observe the effect of the SBES. %

We generated a test environment in simulation with multiple obstacles. The robot's task was to navigate to five randomly set consecutive waypoints.
We set all waypoints at the same depth, as typical navigation with an MBS involves the robot first arriving to the target depth and then navigating along the 2D plane.

\fig{fig:Waypoint_tests_trajectories} shows the trajectories of the three navigation methods and \tab{Quantitative_Analysis_Waypoints} reports the quantitative results measured in terms of traveling time and success ratio. 
Our proposed system with inexpensive monocular camera and SBES achieved the highest navigation efficiency with comparable safety to $\textrm{Bug2}$ with MBS. While the $\textrm{Bug2}$ trajectory appeared not to be affected by noise, it spent the longest navigation time especially when moving along the obstacles. 
Note the echo-sounder played a fundamental role in safe navigation. If the echo-sounder was excluded, the model relied solely on relative monocular image depth estimation to detect surrounding obstacles. As a result, at times the chosen action might be conservative, leading to sub-optimal paths in terms of distance, or too aggressive, increasing the likelihood of collision.  

\subsection{Ablation Study with Transferability Tests} \label{Ablation Study with Transferability Tests}
\vspace{-0.5em}
To show the transferability of our proposed model to different environments and visibilities, we performed an ablation study with the same hyper-parameters and protocols, but considering the following combinations of training settings in a simulated underwater environment:
(1) \textbf{\textit{Rand}}: proposed domain randomization, (2) \textbf{\textit{No Rand (Water)}}: fixed underwater visibility (approximately \SI{11}{m}), and (3) \textbf{\textit{No Rand (Air)}}: no underwater features. 
To firstly exhibit the models' generalizability, another simulated environment\footnote{\scriptsize\url{https://github.com/Scrawk/Ceto}} was employed for testing. With different materials, textures, lightings and custom shaders, it had a different visual appearance compared to the training environment. In this environment, the models were tested in three different scenes, constructed to resemble possible underwater obstacles present in the real-world, such as natural structures (Scene1), submerged wrecks (Scene2) and man-made structures (Scene3).

\begin{figure*}[t]
  \centering
  \resizebox{\textwidth}{!}{
  \begin{tabular}{ccccccc}
     \textbf{Scenes} & \multicolumn{2}{c}{\textbf{\SI[detect-weight=true]{8}{m}}} & \multicolumn{2}{c}{\textbf{\SI[detect-weight=true]{12}{m}}} &\multicolumn{2}{c}{\textbf{\SI[detect-weight=true]{20}{m}}} \\ 
     \textbf{Scene1}
     & \includegraphics[height=1in,%
     valign=c]{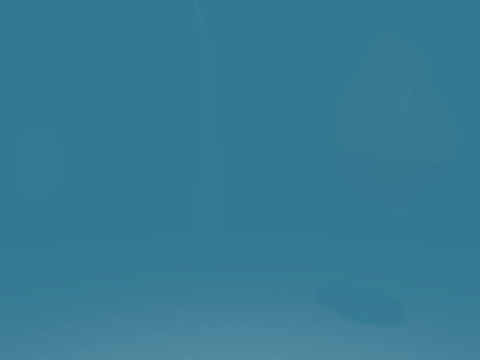} 
     & \includegraphics[height=1.2in,%
     valign=c,trim={3cm 1cm 3cm 3cm},clip]{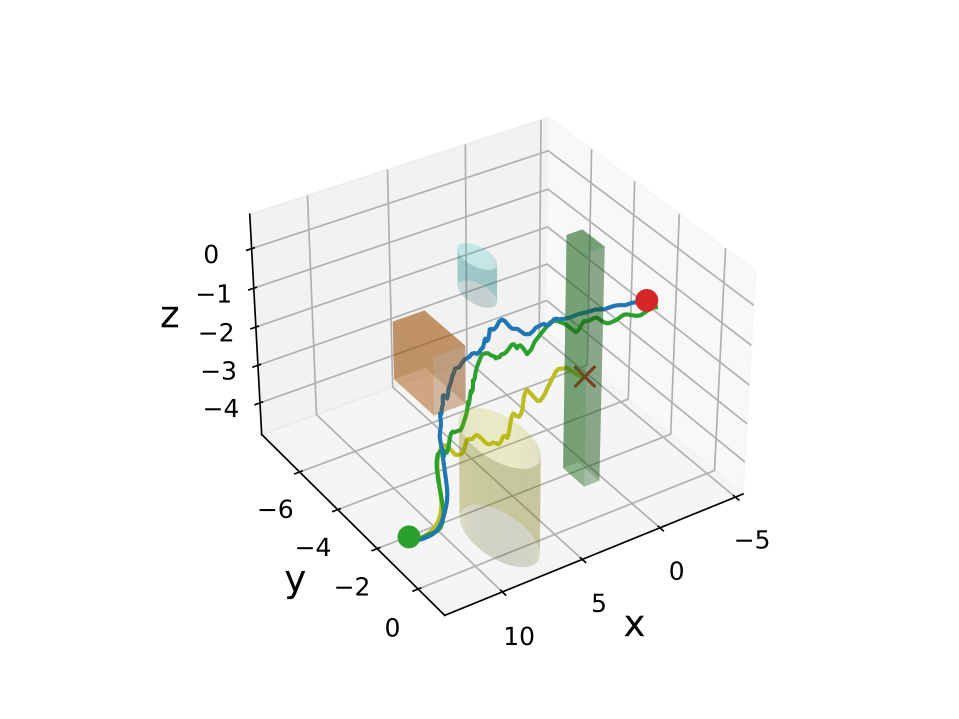}
     & \includegraphics[height=1in,%
     valign=c]{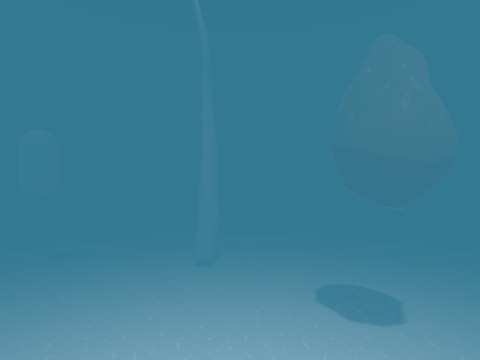} 
     & \includegraphics[height=1.2in,%
     valign=c,trim={3cm 1cm 3cm 3cm},clip]{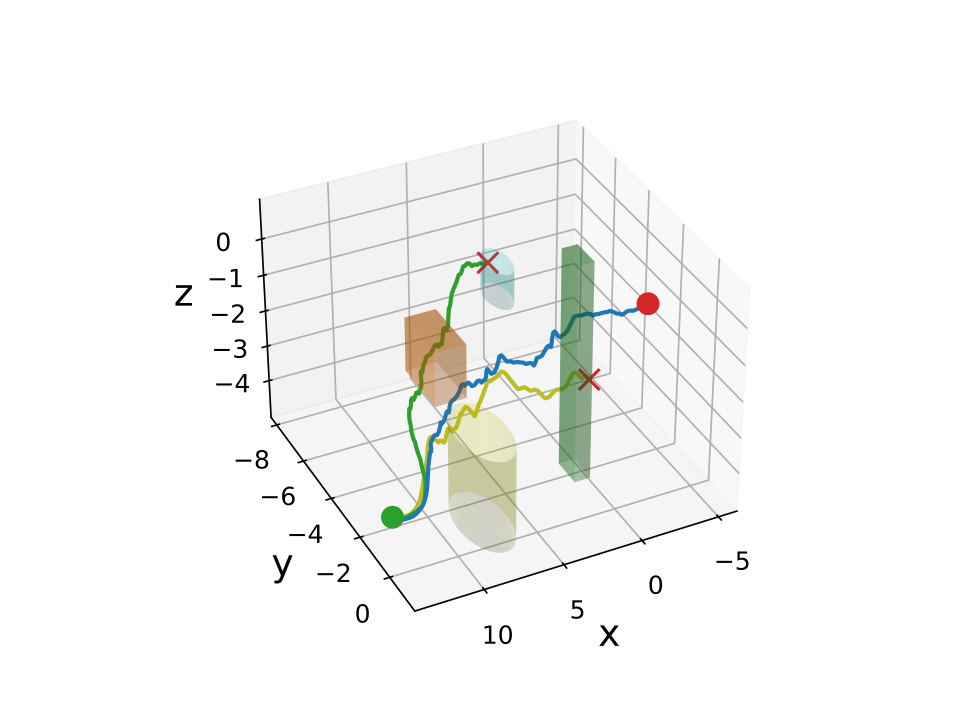} 
     & \includegraphics[height=1in,%
     valign=c]{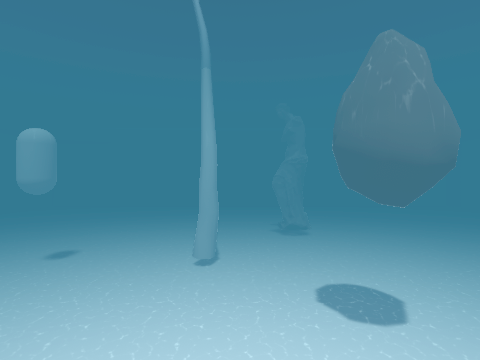} 
     & \includegraphics[height=1.2in,%
     valign=c,trim={3cm 1.5cm 3cm 1.5cm},clip]{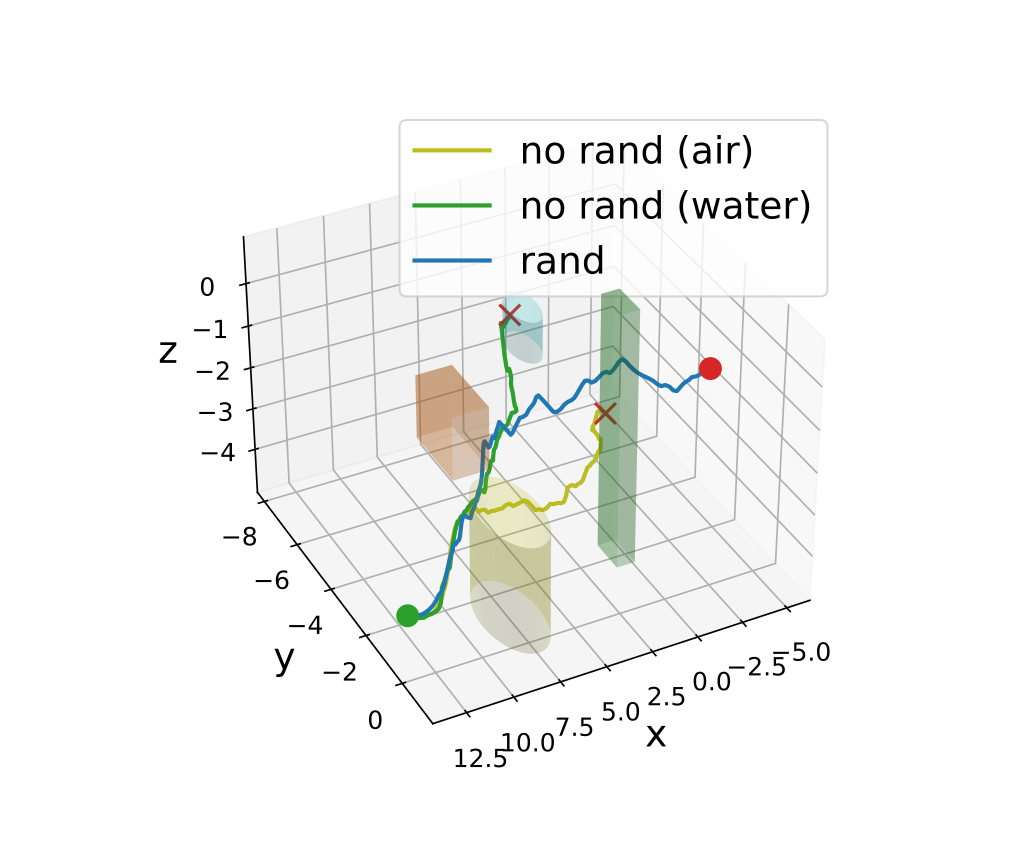}\\ 
     \textbf{Scene2}
     & \includegraphics[height=1in,%
     valign=c]{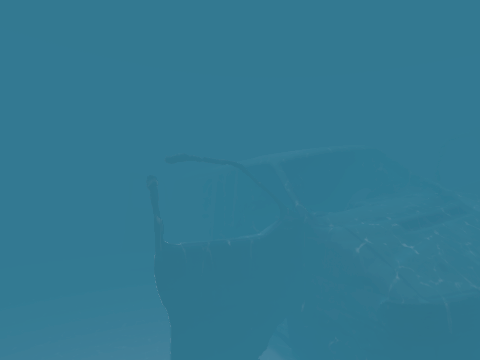} 
     & \includegraphics[height=1.2in, %
     valign=c,trim={3cm 0.8cm 3cm 3cm},clip]{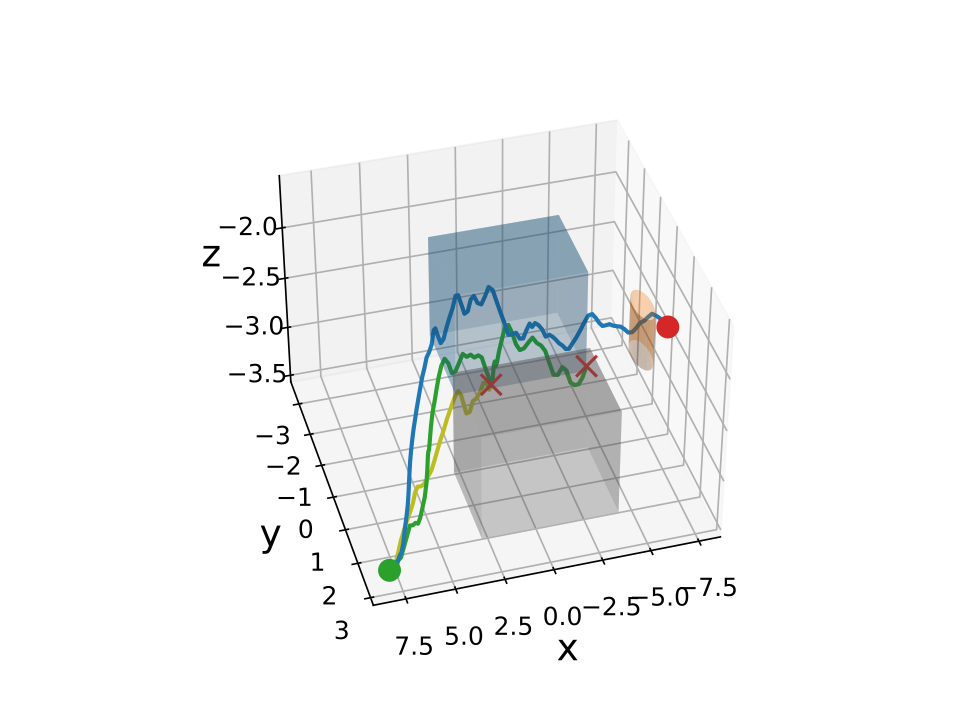} 
     & \includegraphics[height=1in,%
     valign=c]{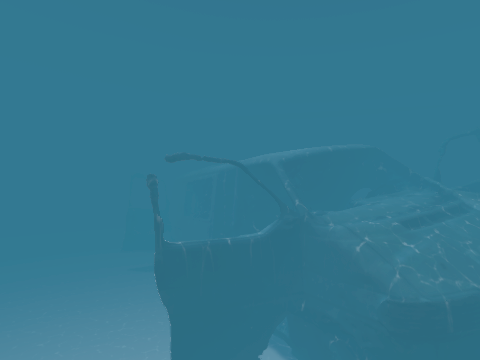} 
     & \includegraphics[height=1.2in, %
     valign=c,trim={3cm 0.8cm 3cm 3cm},clip]{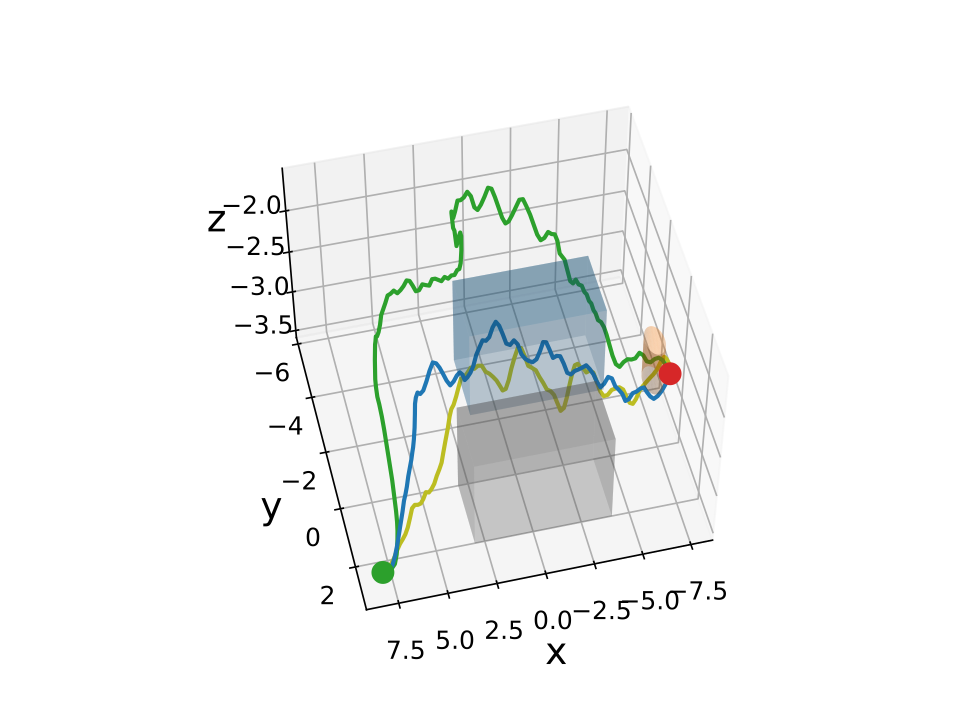} 
     & \includegraphics[height=1in,%
     valign=c]{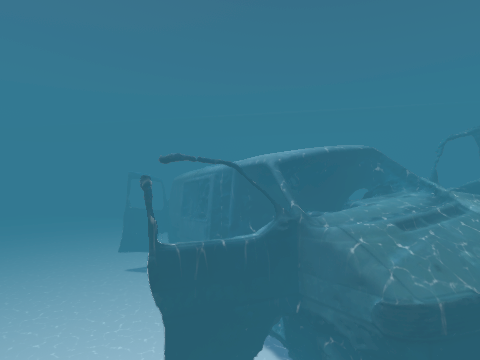} 
     & \includegraphics[height=1.2in, %
     valign=c,trim={3cm 0.8cm 3cm 3.2cm},clip]{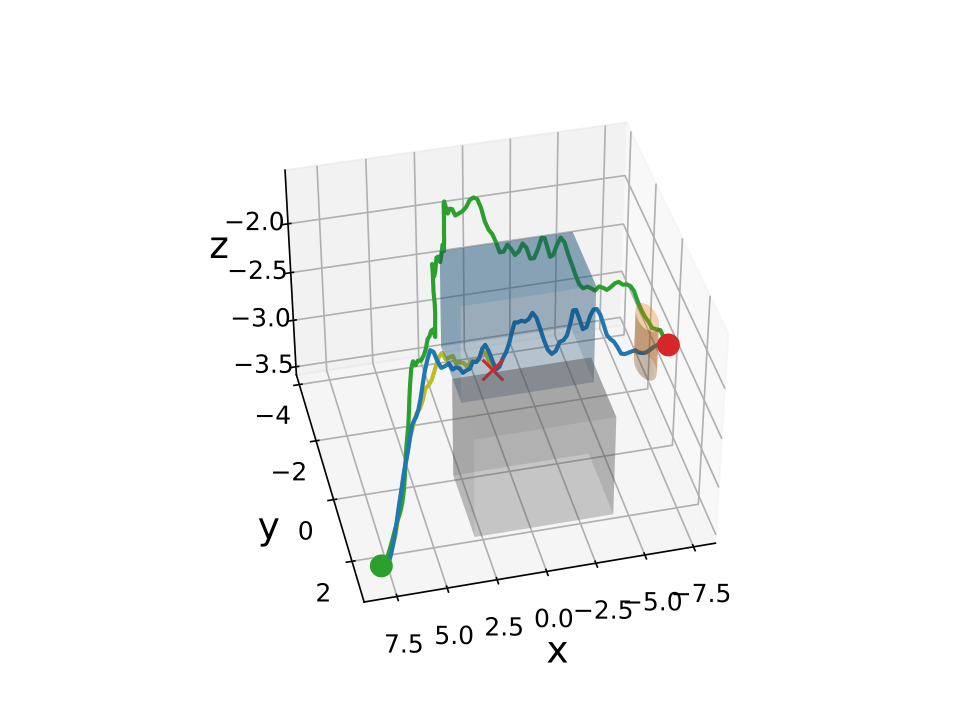} \\ 
     \textbf{Scene3}
     & \includegraphics[height=1in,%
     valign=c]{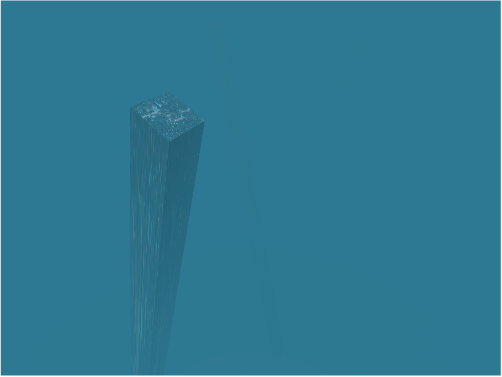} 
     & \includegraphics[height=1.2in, %
     valign=c,trim={4cm 1cm 3cm 3.2cm},clip]{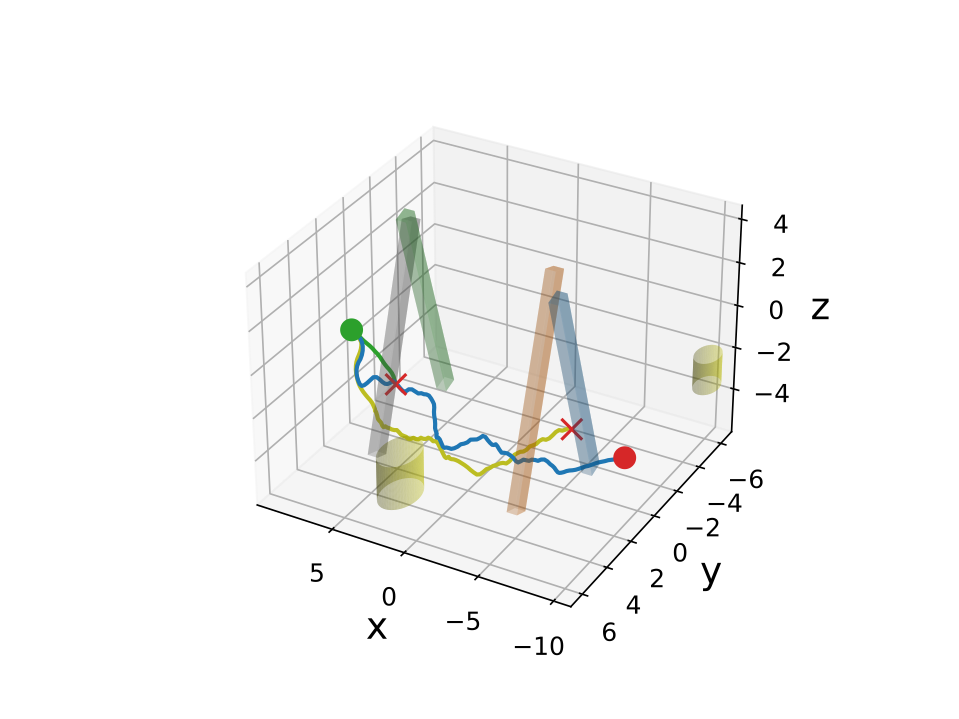}
     & \includegraphics[height=1in,%
     valign=c]{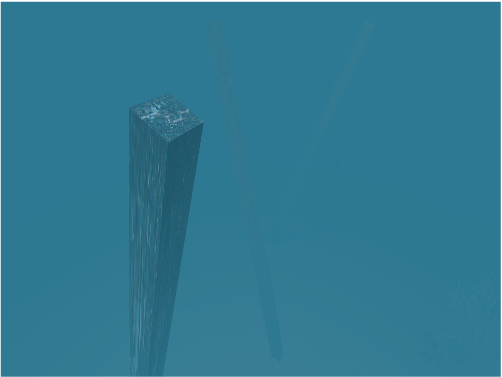} 
     & \includegraphics[height=1.2in,%
     valign=c,trim={4cm 1cm 3cm 3.2cm},clip]{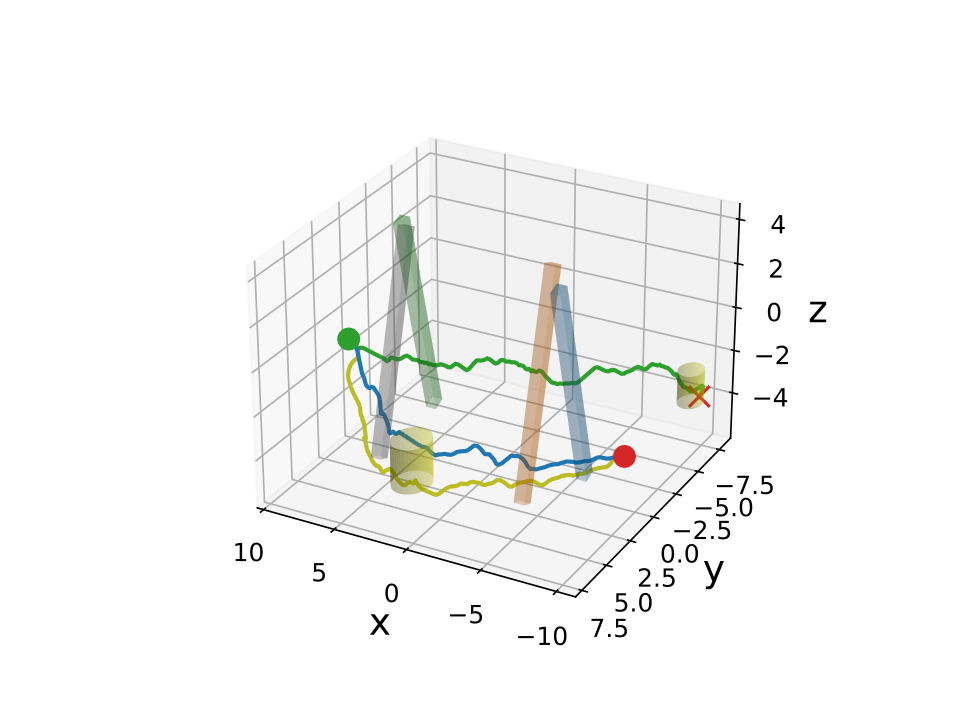}
     & \includegraphics[height=1in,%
     valign=c]{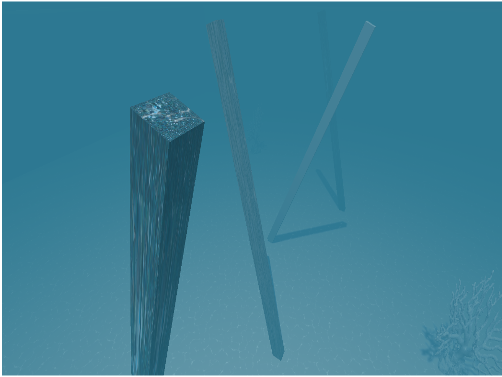} 
     & \includegraphics[height=1.2in,%
     valign=c,trim={4cm 1cm 3cm 3.2cm},clip]{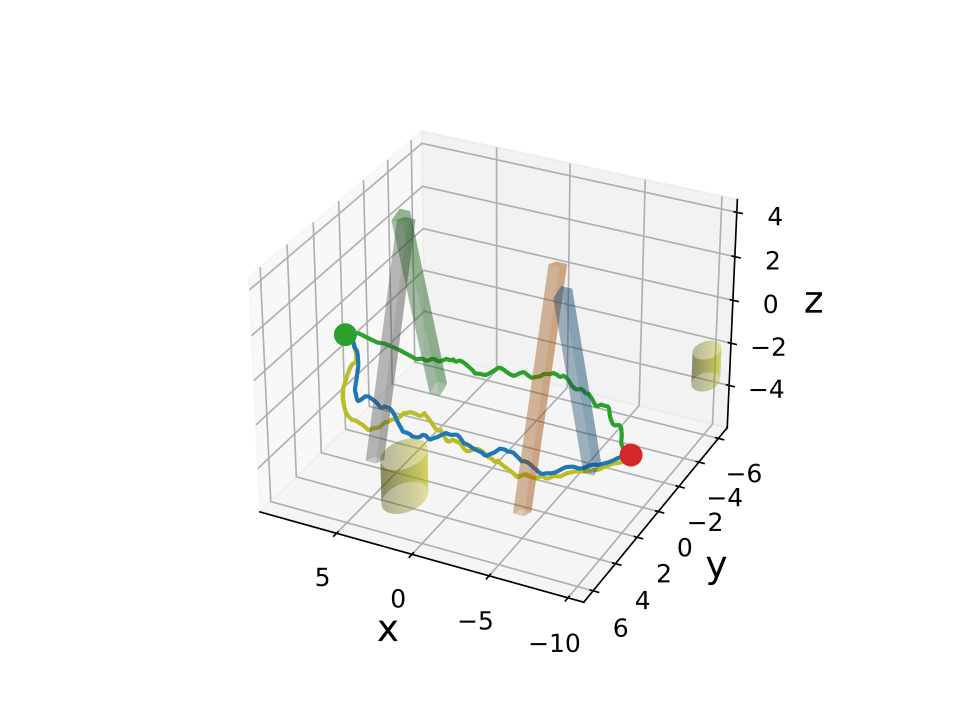} \\ 
  \end{tabular}
  }
  \caption{\textit{Example of Trajectories in Different Scenes with Different Training.} Legend: robot's initial position and goal waypoint (green and red dots); robot collision (red ``X''); obstacles (approximated with polygons in the plots for simplicity).
  }
  \label{figure of 3D trajectories}
  \vspace{-2em}
\end{figure*}

\begin{table*}
\centering
\caption{\textit{Quantitative Results for Transferability Tests.} 10 runs for the three models in three scenes with different visual conditions. %
Note: N/A means the method failed to reach the goal during the runs and bold means the best result.
}
\label{Transferability Comparison Tests}
\resizebox{\textwidth}{!}{
    \begin{tabular}{ccccccccccc}
        \toprule
          \multirow{2}*{Method} & & \multicolumn{3}{c}{$\textrm{Scene1}$} & \multicolumn{3}{c}{$\textrm{Scene2}$} & \multicolumn{3}{c}{$\textrm{Scene3}$} \\\cline{3-5}\cline{6-8}\cline{9-11}
           & & Blurry & Medium & Clear & Blurry & Medium & Clear & Blurry & Medium & Clear\\
        \midrule
         & reward & 5.74 $\pm$ 2.17 & 6.5 $\pm$ 5.95 & 28.14 $\pm$ 2.85 & 0.43 $\pm$ 2.26 &
         10.93 $\pm$ 11.31 & 12.05 $\pm$ 8.92 & 24.64 $\pm$ 10.19 & 20.58 $\pm$ 13.7
         & 29.18 $\pm$ 8.01
        \\
        No\ Rand\ (Air) & success & 0\% & 10\% & \textbf{100\%} & 0\% &
         40\% & 50\% & 70\% & 60\% & 90\%
         \\ 
          & trav. time & N/A & 70.0 & 67.2 $\pm$ 0.84 & N/A &
         \textbf{53.12 $\pm$ 0.65} & 55.2 $\pm$ 2.84 & 63.29 $\pm$ 0.88 & 66.5 $\pm$ 4.53
         & 66.11 $\pm$ 1.07
         \\ \hline
         & reward & \textbf{25.27 $\pm$ 8.42} & 18.35 $\pm$ 11.18 & 13.46 $\pm$ 14.51 & 2.19 $\pm$ 1.78 &
         -1.58 $\pm$ 5.94 & 15.04 $\pm$ 10.6 & 18.03 $\pm$ 11.32 & 30.14 $\pm$ 7.5
         & 29.42 $\pm$ 3.27
         \\ 
         No\ Rand\ (Water) & success & \textbf{90\%} & 90\% & 40\% & 0\% &
         10\% & 70\% & 60\% & \textbf{90\%} & \textbf{100\% }
         \\
         & trav. time & 70.5 $\pm$ 4.93 & 88.17 $\pm$ 18.36 & 69.25 $\pm$ 1.35 & N/A &
         115.0 & 59.79 $\pm$ 8.25 & 71.42 $\pm$ 6.9 & 73.39 $\pm$ 2.63
         & 65.35 $\pm$ 0.78
         \\ \hline

         & reward & 24.66 $\pm$ 9.3 & \textbf{28.39 $\pm$ 2.26} & \textbf{29.56 $\pm$ 2.58} & \textbf{21.68 $\pm$ 9.61} &
         \textbf{23.36 $\pm$ 7.49} & \textbf{24.86 $\pm$ 2.92} & \textbf{29.17 $\pm$ 11.34} & \textbf{30.26 $\pm$ 9.25}
         & \textbf{36.26 $\pm$ 0.83}
         \\
         Rand & success & \textbf{90\%} & \textbf{100\%} & \textbf{100\%} & \textbf{80\%} &
         \textbf{90\%} & \textbf{100\%} & \textbf{80\%} & \textbf{90\%} & \textbf{100\%}
         \\
         & trav. time & \textbf{67.56 $\pm$ 0.44} & \textbf{68.45 $\pm$ 0.72} & \textbf{67.05 $\pm$ 1.27} & \textbf{52.0 $\pm$ 0.35} &
         53.44 $\pm$ 1.23 & \textbf{50.75 $\pm$ 0.46} & \textbf{60.75 $\pm$ 0.56} & \textbf{62.56 $\pm$ 0.98}
         & \textbf{61.05 $\pm$ 0.57}
         \\
        \bottomrule
    \end{tabular}
}

\end{table*}

We considered three visibility scenarios: blurry, medium, and relatively clear, with maximum visibility ranges of  \SI{8}{m}, \SI{12}{m}, and \SI{20}{m}, respectively. 
\fig{figure of 3D trajectories} shows snapshots of each scene and the resulting trajectories in some sample runs. 

\textbf{Comparison metrics.} 
The following metrics were used to compare the three methods' performances (see \tab{Transferability Comparison Tests}):
\begin{itemize}
  \item [1)] 
  Rewards (higher is better): cumulative reward average and standard deviation over $10$ runs,
  \item [2)]
  Success Ratio (higher is better): number of times the robot reached the goal with no collision over $10$ runs, %
  \item [3)]
  Travel Time (less is better): average and standard deviation traveling time ($s$). Failed runs were not considered.
\end{itemize}

From the results, training with underwater features has the highest gain. Adding domain randomization allows a further increase of the cumulative rewards, success rate, and travel time.
Models trained without randomization did not previously encounter abundant visual conditions, thus explored a limited observation space. Accordingly, they would not be easily applicable to different visibility conditions and are more vulnerable to noise especially in low-visibility environments when depth estimations are inaccurate. Scene3 in particular was challenging with blurry visibility, due to the narrow passage between the logs. 

\subsection{Performance Demonstration in Real-World Environment} \label{Performance Demonstration in Real-World Environment}
We conducted real-world experiments with a BlueROV2 in a swimming pool. 
The robot was equipped with a Sony IMX322LQJ-C camera\footnote{\scriptsize\url{https://www.bluerobotics.com/store/sensors-sonars-cameras/cameras/cam-usb-low-light-r1/}} 
with a resolution of 5 MP, a horizontal and vertical FOV of \ang{80} and \ang{64}.
The fixed SBES has a \ang{30} beam width and a maximum range set to \SI{4}{m}.
The (noisy) robot's pose was provided by an on-board compass, a water-pressure sensor to recover water depth, and a short baseline acoustic positioning system (SBL)\footnote{\scriptsize\url{https://waterlinked.github.io/explorer-kit/introduction/}}. %
A \SI{2.8}{GHz} Intel i7 laptop with Nvidia Quadro M1200 was used for running the inference network through the Robot Operating System (ROS). For real-time inference, DPT was replaced with its computationally less expensive counterpart MiDaS~\cite{ranftl2019towards} as our depth prediction network -- about 0.08 seconds per inference.
\begin{figure}
  \centering
  \resizebox{\textwidth}{!}{
  \begin{tabular}{c c c c}
     \includegraphics[height=.5in,trim={.3cm .4cm .3cm .6cm}, clip,%
     valign=c]{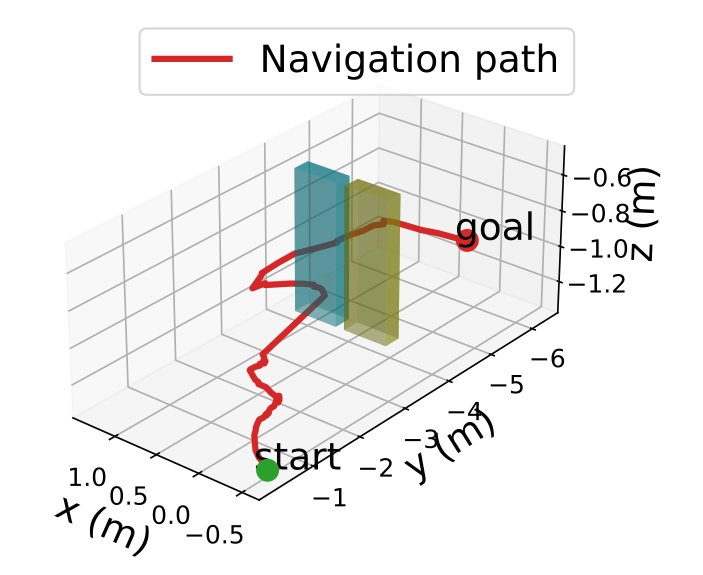}
     & \includegraphics[height=.5in,%
     trim={.3cm .4cm .3cm .6cm}, clip,valign=c]{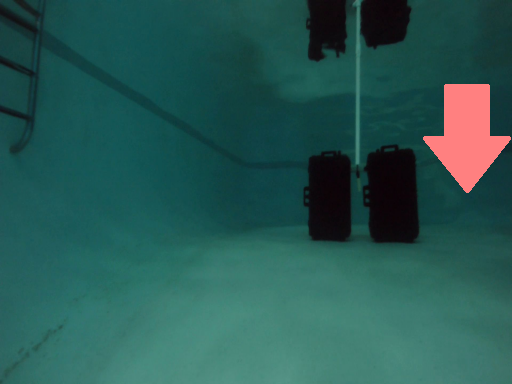} 
     \includegraphics[height=.5in,trim={3cm .4cm 1.5cm 3cm}, clip,%
     valign=c]{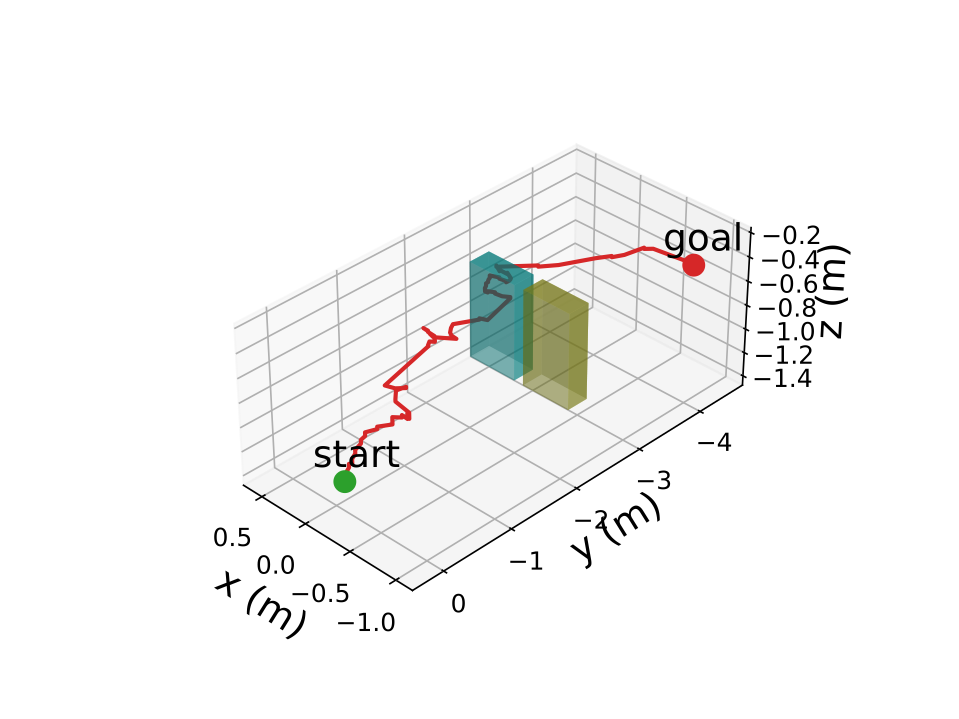}
      & \includegraphics[height=.5in,%
      valign=c]{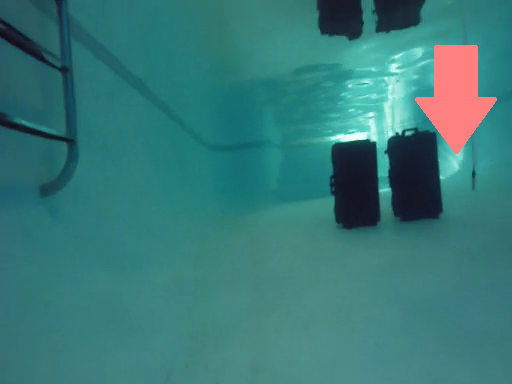} \\
     \includegraphics[height=.4in,trim={.3cm .4cm .3cm .6cm}, clip,%
     valign=c]{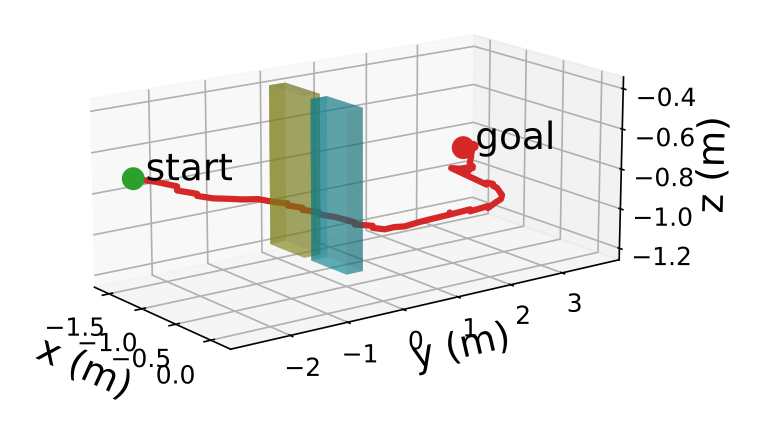}
      & \includegraphics[height=.5in,trim={.3cm .4cm .3cm .6cm}, clip,%
      valign=c]{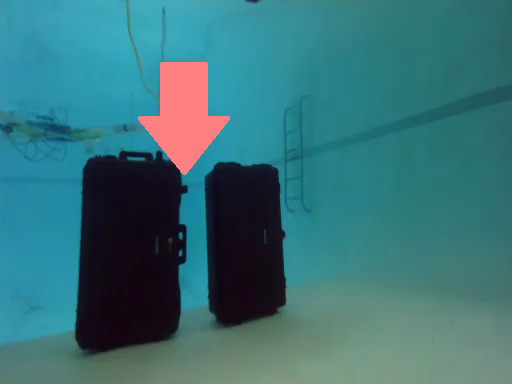} 
     \includegraphics[height=.5in,trim={.3cm .4cm .3cm .6cm}, clip,height=.6in,%
     valign=c]{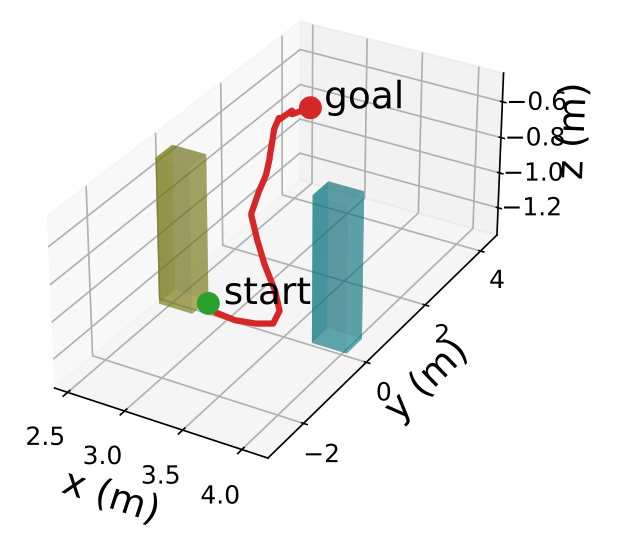}
      & \includegraphics[height=.5in,%
      valign=c]{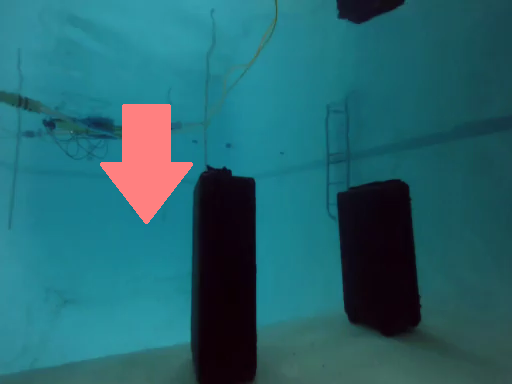}
  \end{tabular}
  }
  \vspace{-1em}
  \caption{\textit{Pool Experiment.} Navigation trajectories with localization noise smoothing (legend: Start and goal green and red dots; obstacles, cuboids) and  images from the robot's camera. Red arrows point to the approximate goal locations behind the boxes.}
  \label{fig:table_of_paths_and_images}
  \vspace{-1em}
\end{figure}

The swimming pool was about \SI{20}{m} by \SI{7}{m} in size with a shallow (\SI{1}{m}) and deep (\SI{3}{m}) end, and a slope in the middle. Two black boxes (approximate size: 0.8  x 0.5 x 0.3 m
were placed in two different configurations: side by side as a large obstacle and with a \SI{1}{m} separation to create a channel.

Resulting paths and reference images are shown in \fig{fig:table_of_paths_and_images}.
Our proposed navigation approach successfully drove the BlueROV2 to different 3D waypoints, avoiding obstacles by going around, above, or through a channel (see \fig{fig:table_of_paths_and_images}).
We observed that the SBL provided noisier position information compared to in simulation -- at times the robot's location jumped up to a meter. %
While the noise affected the calculation of the relative position to the goal, our approach does not depend on the absolute robot location to infer obstacle distance, so the robot was able to avoid obstacles.

\vspace{-0.5em}
\subsection{Action Prediction from Static Underwater Images}
\begin{figure}[t]
  \centering
  \begingroup
\renewcommand{\arraystretch}{4} %
\resizebox{\textwidth}{!}{
  \begin{tabular}{c c c c c c c}
     \includegraphics[width=.29\columnwidth, valign=c]{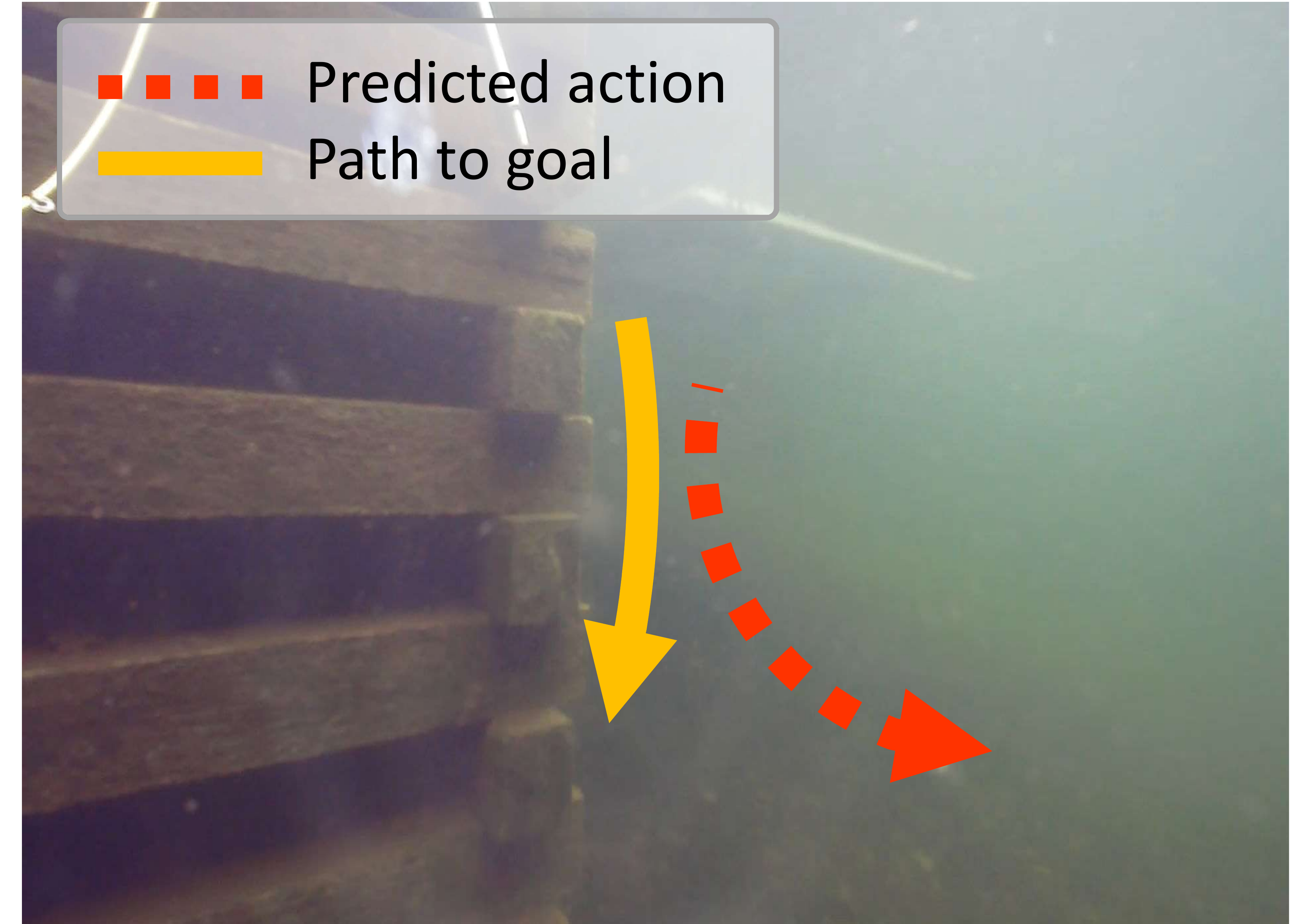}
     & \includegraphics[width=.29\columnwidth, valign=c]{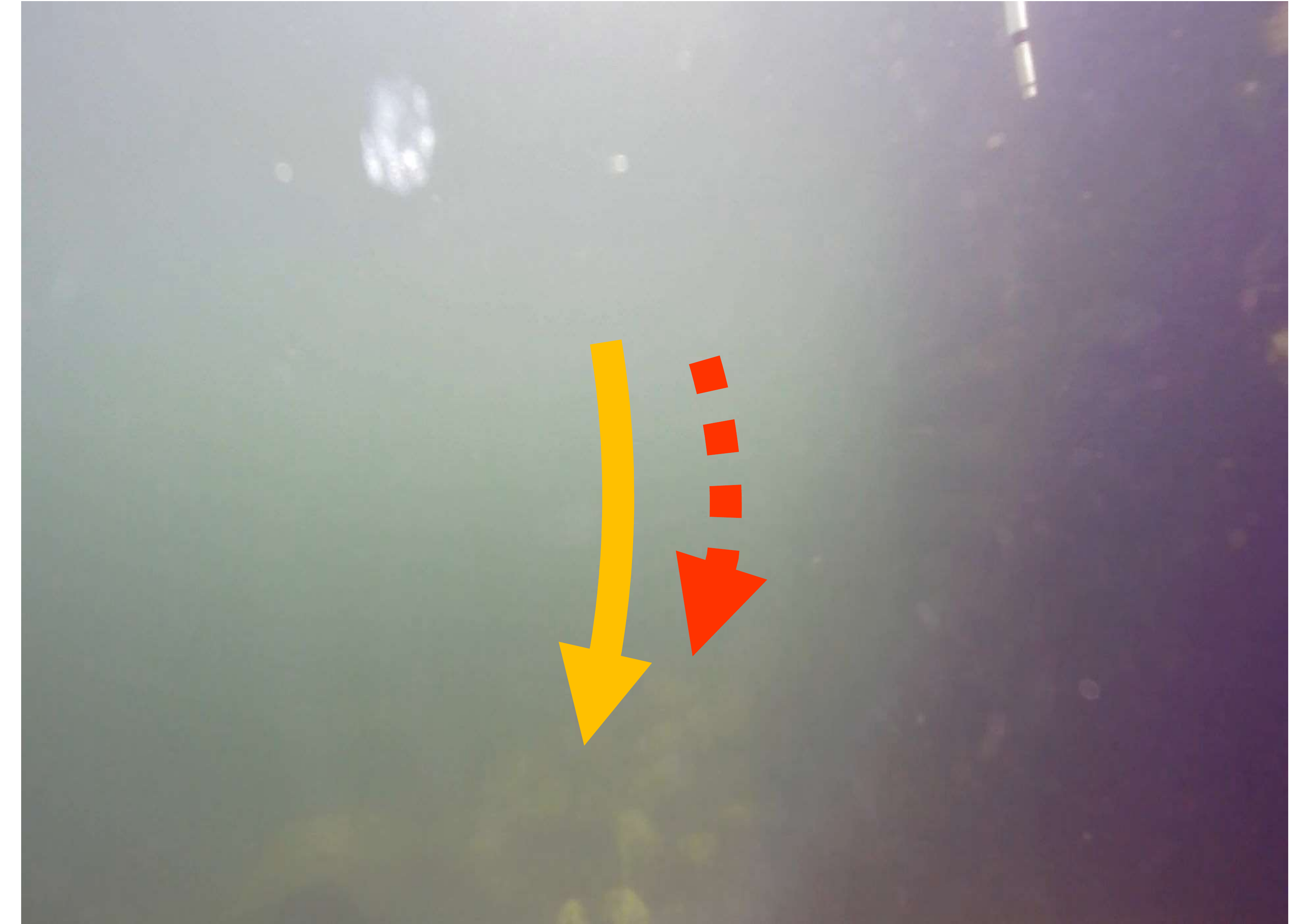}
     & \includegraphics[width=.29\columnwidth, valign=c]{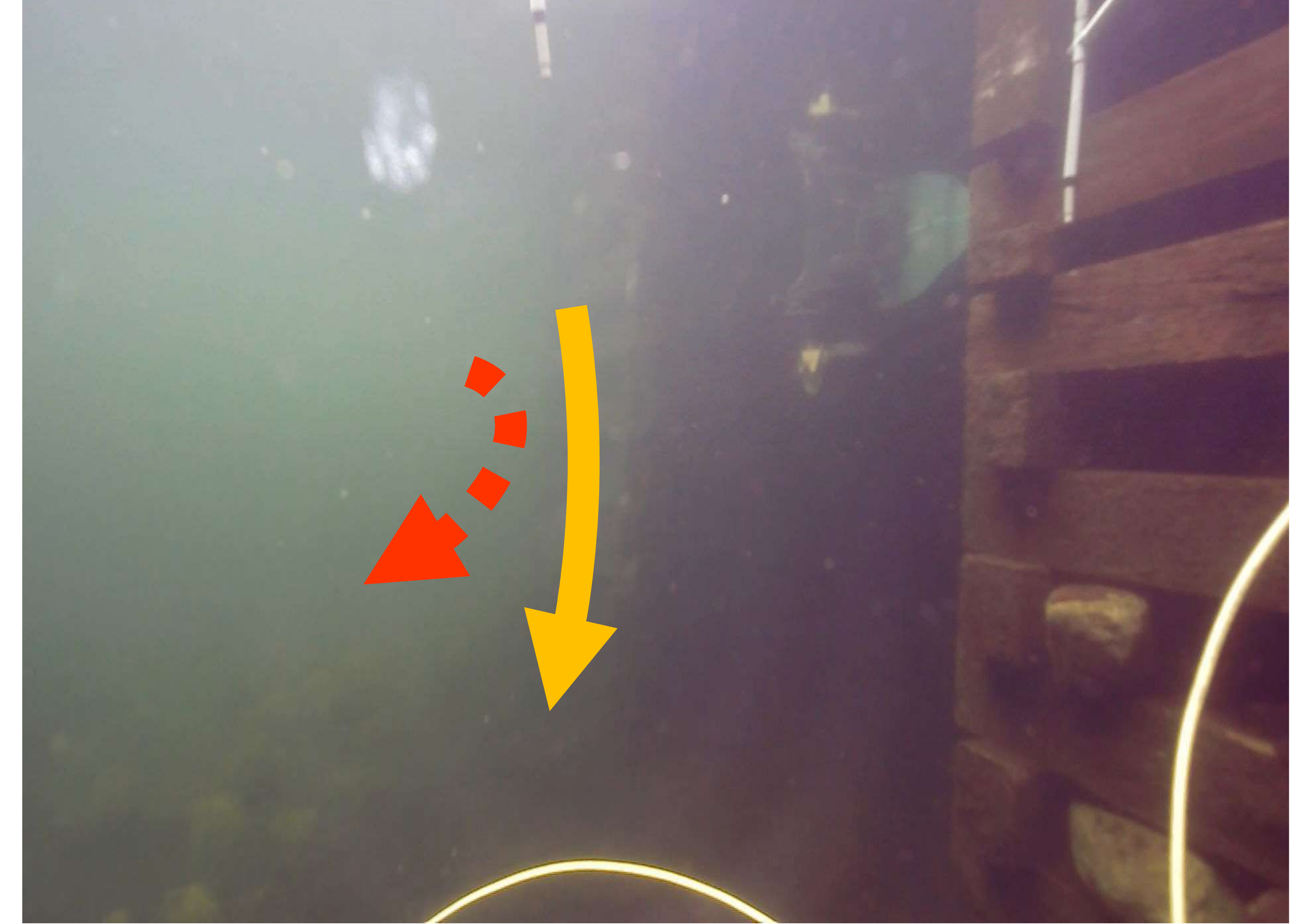} 
     \includegraphics[width=.27\columnwidth, valign=c]{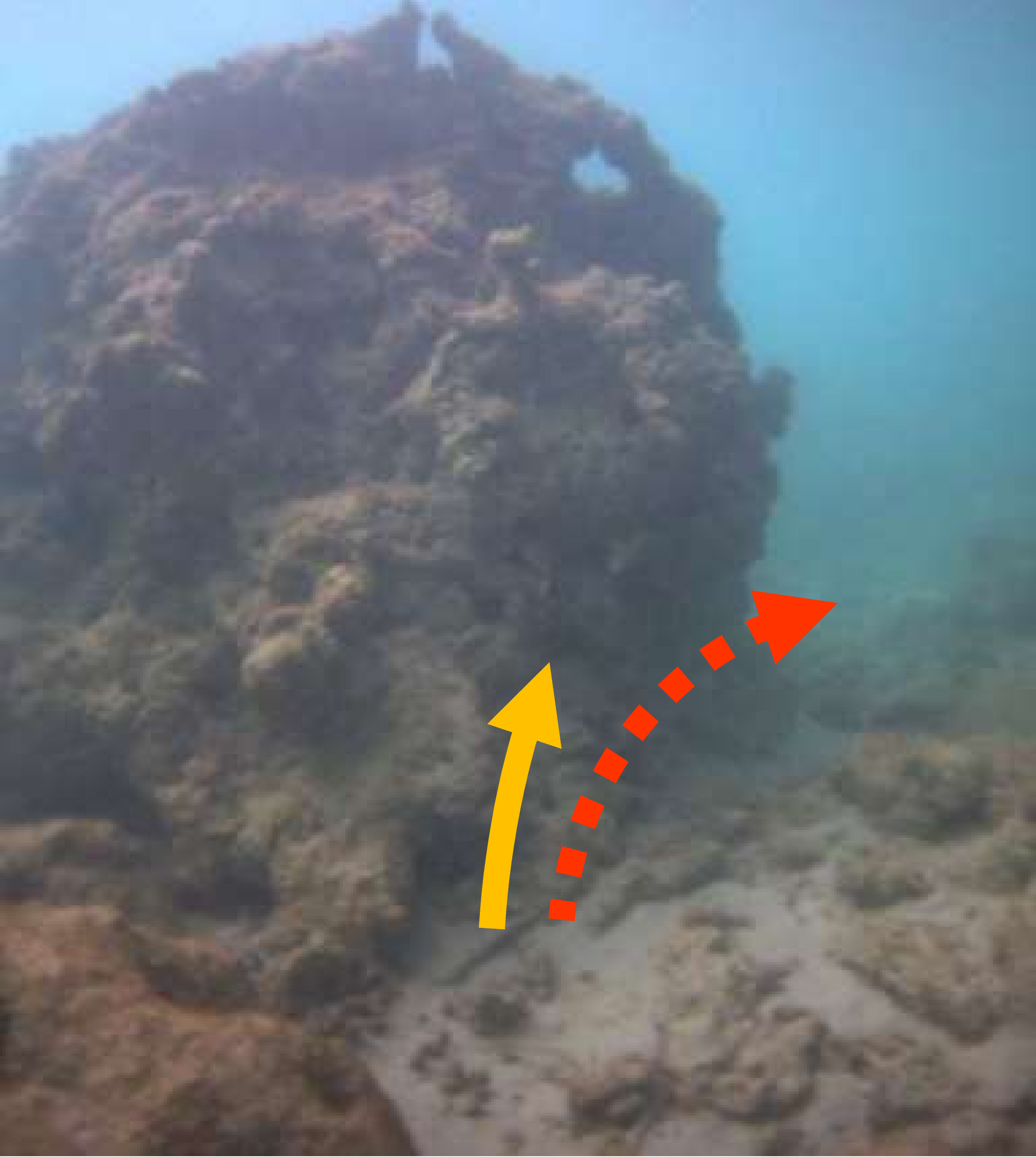}
     & \includegraphics[width=.27\columnwidth, valign=c]{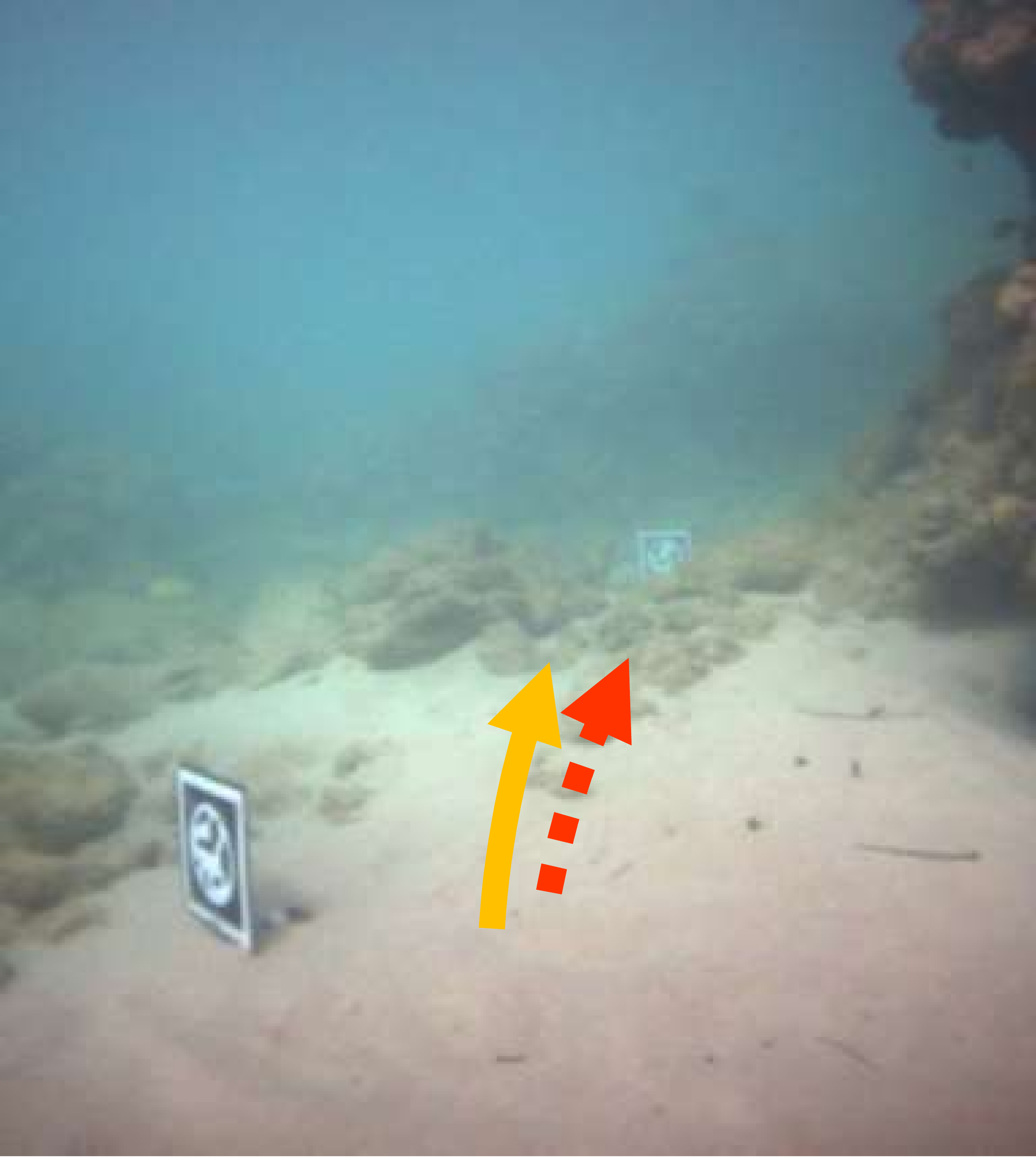}
     & \includegraphics[width=.27\columnwidth, valign=c]{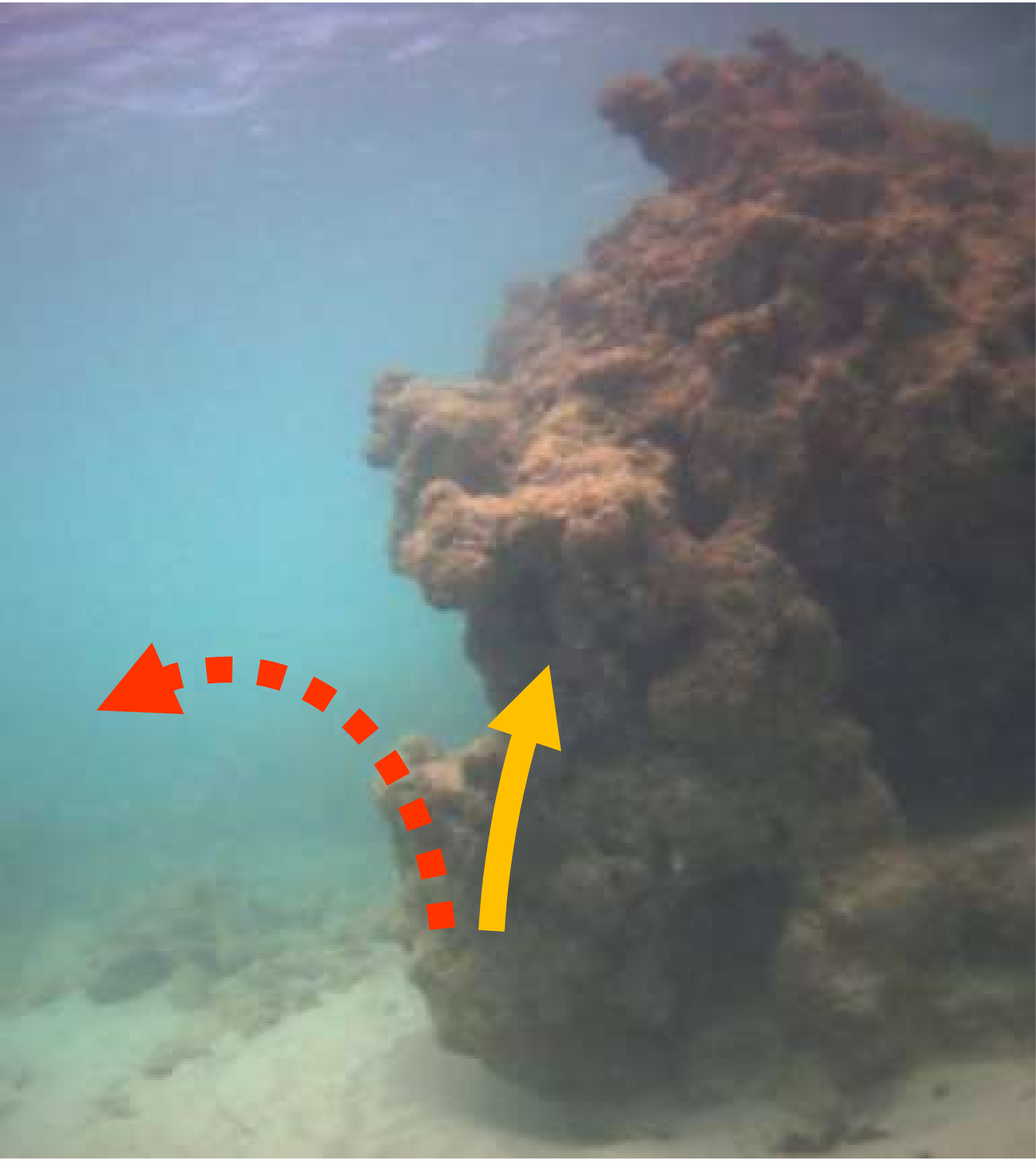} 
      \\\hline
     \includegraphics[width=.29\columnwidth, valign=c]{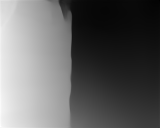}
     & \includegraphics[width=.29\columnwidth, valign=c]{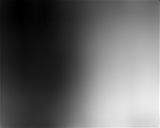}
     & \includegraphics[width=.29\columnwidth, valign=c]{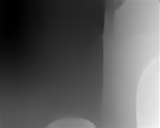}
     \includegraphics[width=.29\columnwidth, valign=c]{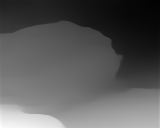}
     & \includegraphics[width=.29\columnwidth, valign=c]{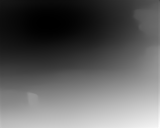}
     & \includegraphics[width=.29\columnwidth, valign=c]{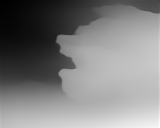}
  \end{tabular}
  }
  \endgroup
  \vspace{-1em}
  \caption{\textit{Single Image Action and Depth Prediction.} 1st row: images from Lake Sunapee and Caribbean Sea. 2nd row: their respective depth predictions. Direction and magnitude of the action predicted (red arrow); approximate goal location (yellow arrow).}
  \vspace{-2em}
  \label{fig:table_of_oceanic_image_depth_estimation}
  \end{figure}

We also tested joint image and SBES reading data from past field trials (in the Caribbean Sea and lake) as input to our model for action prediction.
\fig{fig:table_of_oceanic_image_depth_estimation} shows a sample of such images with corresponding depth predictions, locations of the goal, and predicted actions.
As expected, with obstacles nearby the predicted action prioritized obstacle avoidance, steering the robot away, otherwise, the action's direction pointed towards the goal. This qualitative test demonstrates our model's generalizability to real-world applications.
\section{Conclusion and Future Work}\label{sec:conclusion}

We presented the first 3D map-less underwater navigation approach, based on Proximal Policy Optimization Network (PPO) and domain randomization, for low-cost underwater robots with a monocular camera and a fixed single-beam echo-sounder. By choosing deep reinforcement learning over classic methods, we were able to address the intrinsic challenges of seamless underwater navigation (e.g., lack of low-cost efficiency sensor and difficulty in generating a map given noisy positioning and perception data). We validated our approach with several comparisons and ablation studies in different simulated environments, as well as real-world validation in a swimming pool and  with static underwater images. Results showed that the robot is able to navigate to arbitrary 3D goals while avoiding obstacles inferred from estimated depth images and sonar readings.

In the future, we will investigate explicit sensor fusion of camera and SBES data to achieve better depth prediction with absolute scale, e.g. early fusion~\cite{roznere2020iros}, as well as controller and SBL data. In addition, we will consider the generation of more complex environments, other real-world experiments, and the design of integrated models for different sensor configurations (e.g., stereo cameras) and dynamic models to adapt our method to heterogeneous underwater robots.

{
\footnotesize
\vspace{-1em}
\section*{Acknowledgments}
\vspace{-1em}
We thank Devin Balkcom for access to the pool for experiments, and Bo Zhu, Mary Flanagan, and Sukdith Punjasthitkul for GPU access. This work is supported in part by the Burke Research Initiation Award and NSF CNS-1919647, 2024541, 2144624, OIA-1923004. 
}

\bibliographystyle{IEEEtran}
\bibliography{IEEEabrv,references}

\end{document}